%% file: main.tex
\documentclass[manuscript,screen]{acmart}
\usepackage{multirow}
\usepackage{caption}
\usepackage{subcaption}
\usepackage{enumitem}
\usepackage{booktabs}
\usepackage{balance}
\usepackage[most]{tcolorbox}
\usepackage{colortbl}
\usepackage{dsfont}
\usepackage[linesnumbered,lined]{algorithm2e}
\usepackage{algpseudocode}
\usepackage{xcolor}

\setlist[itemize]{leftmargin=*}

\AtBeginDocument{%
  }

\setcopyright{acmlicensed}
\copyrightyear{2025}
\acmYear{2025}
\acmDOI{XXXXXXX.XXXXXXX}
\acmJournal{TSAS}
\acmISBN{978-1-4503-XXXX-X/18/06}

\begin{document}

\title[{\textsc{TrajLearn}: Trajectory Prediction Learning using Deep Generative Models}]{\textsc{TrajLearn}: Trajectory Prediction Learning using Deep Generative Models}

\author{Amirhossein Nadiri}
\orcid{0000-0003-4112-2138}
\affiliation{%
  \institution{York University}
  \city{Toronto}    
  \state{Ontario}
  \country{Canada}}
\email{anadiri@yorku.ca}

\author{Jing Li}
\orcid{0000-0002-8913-1159}
\affiliation{%
  \institution{York University}
  \city{Toronto}
  \state{Ontario}
  \country{Canada}}
\email{jliellen@yorku.ca}

\author{Ali Faraji}
\orcid{0000-0002-2439-8493}
\affiliation{%
  \institution{York University}
  \city{Toronto}
  \state{Ontario}
  \country{Canada}}
\email{faraji@yorku.ca}

\author{Ghadeer Abuoda}
\orcid{0000-0001-6501-6175}
\affiliation{%
  \institution{York University}
  \city{Toronto}
  \state{Ontario}
  \country{Canada}}
\email{gha@yorku.ca}

\author{Manos Papagelis}
\orcid{0000-0003-0138-2541}
\affiliation{%
  \institution{York University}
  \city{Toronto}
  \state{Ontario}
  \country{Canada}}
\email{papaggel@eecs.yorku.ca}

\renewcommand{\shortauthors}{Nadiri et al.}

\begin{abstract}
\input{abstract}
\end{abstract}



\received{27 December 2024}
\received[revised]{29 March 2025}
\received[accepted]{2 April 2025}
\maketitle

\input{introduction}
\input{problem}
\input{higher-order-trajectory-representations}
\input{trajectory-prediction-learning}
\input{experiment}
\input{optimization}
\input{related}
\input{ethics}
\input{conclusion}

\begin{acks}
\input{acknowledgments}
\end{acks}

\clearpage
\bibliographystyle{ACM-Reference-Format}
\bibliography{refs}

\end{document}

%% file: abstract.tex
Trajectory prediction aims to estimate an entity's future path using its current position and historical movement data, benefiting fields like autonomous navigation, robotics, and human movement analytics. Deep learning approaches have become key in this area, utilizing large-scale trajectory datasets to model movement patterns, but face challenges in managing complex spatial dependencies and adapting to dynamic environments. 
To address these challenges, we introduce \textsc{TrajLearn}, a novel model for trajectory prediction that leverages generative modeling of higher-order mobility flows based on hexagonal spatial representation. \textsc{TrajLearn} predicts the next $k$ steps by integrating a customized beam search for exploring multiple potential paths while maintaining spatial continuity. 
We conducted a rigorous evaluation of \textsc{TrajLearn}, benchmarking it against leading state-of-the-art approaches and meaningful baselines. The results indicate that \textsc{TrajLearn} achieves significant performance gains, with improvements of up to $\sim$40\% across multiple real-world trajectory datasets. 
In addition, we evaluated different prediction horizons (i.e., various values of $k$), conducted resolution sensitivity analysis, and performed ablation studies to assess the impact of key model components.
Furthermore, we developed a novel algorithm to generate mixed-resolution maps by hierarchically subdividing hexagonal regions into finer segments within a specified observation area. This approach supports {\em selective detailing}, applying finer resolution to areas of interest or high activity (e.g., urban centers) while using coarser resolution for less significant regions (e.g., rural or uninhabited areas), effectively reducing data storage requirements and computational overhead.
We promote reproducibility and adaptability by offering complete code, data, and detailed documentation with flexible configuration options for various applications.

%% file: introduction.tex
\section{Introduction}\label{sec:intro}

\begin{figure}[ht]
    \centering
    \includegraphics[width=0.97\textwidth,trim={0 68pt 0 0},clip=true]{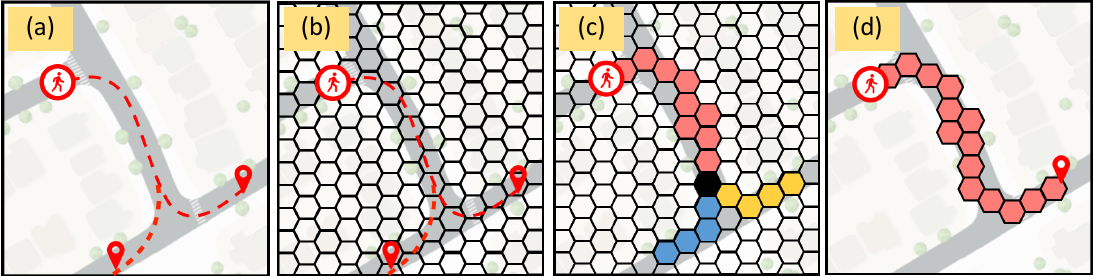}
    \captionsetup{font=small}
    \caption{Illustrative example of the trajectory prediction problem using higher-order spatial representations (hexagons); ({\em a}) there are two \textcolor{red}{potential} trajectories for the pedestrian, ({\em b}) trajectories are represented on a hexagon-based tessellated map, ({\em c}) given the historical data (\textcolor{red}{red}) and the current location (black), two trajectories are predicted (\textcolor{blue}{blue} and \textcolor{orange}{orange}), ({\em d}) the actual trajectory followed.}
    \Description{A multi-panel figure illustrating trajectory prediction using a hexagon-based tessellated map. In panel (a), a pedestrian faces two potential trajectories. Panel (b) shows these trajectories mapped onto a grid of hexagons. In panel (c), historical data (in red) and the current position (in black) are used to predict two future trajectories, highlighted in blue and orange. Panel (d) displays the actual path followed.}
    \label{fig:teaser}
\end{figure}

\smallskip\noindent\textbf{Motivation \& Problem of Interest.} The development of tracking and geolocation technology has facilitated the collection of large-scale mobility data, encompassing both objects and individuals \cite{wang2021analytics, sakr2022analytics}. Mining interesting patterns in mobility data is of increased research and development interest due to a wide range of practical applications. Technical problems in the area include trajectory classification, clustering, prediction, simplification, and anomaly detection (see~\cite{zheng2015trajectory, alturi2018spatiotemporal, hamdi2022spatiotemporal} for comprehensive surveys). In this research, we focus on the {\em trajectory prediction problem}, which refers to the task of predicting the future path or trajectory of an object (or individual) based on its current state and historical data. Efficient methods for trajectory prediction are highly desirable in various domains and applications, including transportation systems, human mobility studies, autonomous vehicles, robotics, and more.

\smallskip\noindent\textbf{The State of the Art \& Limitations.}  
First attempts to address the problem considered statistical methods, such as matrix factorization \cite{cheng2012fused, li2015rank, lian2014geomf} and Markov chain \cite{cheng2013you, mathew2012predicting, shi2019semantics}. However, these methods frequently encounter challenges in capturing trajectories' intricate sequential and periodic characteristics.
Recent progress in Deep Learning (DL) has led to the emergence of deep neural models explicitly tailored to capture the sequential characteristics inherent in trajectories. Notably, approaches centered around Recurrent Neural Networks (RNNs) have exhibited promising results~\cite{liu2016predicting, feng2018deepmove}. However, despite their favorable performance, these models encounter challenges when confronted with sparse and imprecise trajectory data~\cite{alsaeed2023trajectory, isufaj2023gti}. Additionally, they often demand substantial quantities of meticulously labeled training data, a resource-intensive and time-consuming endeavor. Furthermore, there is a risk of overfitting the training dataset, resulting in suboptimal generalization capabilities when faced with unseen data. 
A {\em pertinent} but {\em distinct} problem from the one addressed in this paper is the problem of predicting the Next Point of Interest (POI). It is \textbf{pertinent} as it involves predicting future locations, specifically the POIs that a user or object is likely to visit next in a sequence of discrete locations, such as restaurants, shops, tourist attractions, gas stations, etc. Numerous studies have extensively explored this topic and proposed efficient methods like ST-RNN~\cite{liu2016predicting}, ST-LSTM~\cite{kong2018hst}, STAN~\cite{luo2021stan} and Graph-flashback~\cite{rao2022graph}, to name a few. These methods typically assume the input data is a history of POI check-ins. Nonetheless, it is \textbf{distinct}, as our problem predicts {\em the entire future path or trajectory} that a user or object will follow based on their historical movement data. 
Certain prior studies explored trajectory prediction using comprehensive GPS log datasets instead of merely relying on POI check-ins~\cite{jiang2018deepurbanmomentum, amichi2021movement, yan2023precln}. However, these studies address specialized versions of the problem and mostly rely on semantic information from the datasets to train their model, which lacks effective generalization to the broader issue at hand.
Due to this distinction, our proposed methods are not directly comparable to methods proposed for the Next POI problem, as explained in the experiments (see section~\ref{sec:exp}). 

\smallskip\noindent \textbf{Our Approach \& Contributions.} To address these limitations, we propose a novel approach that leverages deep generative models 
to accurately predict the future path of a user or an object based on historical data.
Our contributions can be summarized as follows:
\begin{itemize}[leftmargin=6mm]
\item We formalize the trajectory prediction problem as {\em a sequence prediction problem}. Given as input the recent history of a trajectory, represented as a sequence of continuous blocks (hexagons) of a regularly tessellated map, the task is to predict the trajectory's future $k$ continuous blocks.

\item We propose \textsc{TrajLearn}, a trajectory generative model based on the Transformer architecture \cite{vaswani2017attention}. \textsc{TrajLearn} is trained (from scratch) on historical trajectory data provided in the form of higher-order mobility flow data and incorporates a variant of beam search to simultaneously explore multiple candidate paths while respecting spatial constraints for path continuity (see example in Figure \ref{fig:teaser}). \textsc{TrajLearn} novelty contribution lies in its comprehensive approach, combining these elements with a unique constrained beam search guided by spatial relationships.
The method is versatile and can forecast future trajectories at different levels of granularity, enabling diverse levels of analysis and applications. 

\item We design and develop a novel algorithm that generates mixed-resolution maps by hierarchically subdividing hexagonal regions into finer segments within a defined observation area, enhancing adaptability and enabling broader applicability to various trajectory analysis scenarios.
\item We demonstrate empirically that \textsc{TrajLearn} outperforms the state-of-the-art methods and sensible baselines by as much as $\sim$40\%, on various evaluation metrics and diverse real-world trajectory datasets. We also study the parameter sensitivity and model ablation to assess how \textsc{TrajLearn} behaves under various configurations and parameters.

\item We open-source our code and model to encourage reproducibility (see details below).
\end{itemize}

\begin{tcolorbox}[colback=black!5,colframe=black!70, boxrule=.2mm, boxsep=.2mm, top=0.2mm]
\smallskip\noindent\textbf{Ensuring Reproducibility and Adaptability.} We provide comprehensive access to source code and data. We take meticulous steps to ensure that all prerequisites are clearly outlined and accompanied by step-by-step instructions to help users set up and run the provided models smoothly. Additionally, we include thoroughly documented model configurations necessary for both the training and testing phases. To further enhance adaptability, we offer clear and detailed documentation of various configuration options, allowing users to modify the model for application in diverse scenarios and customized use cases.

\smallskip\noindent\textbf{GitHub Repository:} \url{https://github.com/amir-ni/trajectory-prediction}
\end{tcolorbox}

\smallskip\noindent\textbf{Broader Impact}. Accurate trajectory prediction offers numerous benefits across various domains and applications. It enhances safety in autonomous driving \cite{huang2022survey} and maritime navigation \cite {xiao2017maritime, chondrodima2022machine} by reducing collision risks, improves resource management and efficiency in logistics and urban planning, and aids in public transport and real-time traffic management for more efficient systems and reduced congestion \cite{wang2021survey}. Additionally, it enables geospatial analysis in urban environments for optimized traffic infrastructure planning and location-based services and recommendations \cite{zheng2015trajectory}. In the social sciences, our model offers valuable insights into human crowd behavior \cite{sawas2019versatile, sawas2018trajectolizer, sawas2018tensor}, while in epidemiology, it supports the development of mobility-based models for understanding the spread of infectious diseases \cite{chang2021mobility, pechlivanoglou2022epidemic, alix2022mobility, pechlivanoglou2022microscopic, yanin2023optimal, Buckee2021}.

\smallskip\noindent\textbf{Paper Organization.} 
Section~\ref{sec:problem} introduces preliminaries and the problem. Section~\ref{sec:higher-order} presents the rationale for utilizing higher-order mobility flow data, and section~\ref{sec:trajectory-prediction-learning} delves into the specifics of the trajectory prediction model. Section~\ref{sec:exp} presents an empirical evaluation of our approach. Section \ref{sec:hierarchical} introduces the use of hierarchical maps for trajectory prediction in complex urban environments. Section \ref{sec:related-work} provides an overview of related work. Section~\ref{sec:ethics} highlights ethical aspects of trajectory prediction models, and section \ref{sec:conclusions} concludes the paper.

%% file: problem.tex
\section{Preliminaries and the Problem}\label{sec:problem}

This section introduces the notation and preliminaries relevant to our model, followed by a formal definition of the problem of interest. A summary of key notation is provided in Table \ref{tab:notations}.

\subsection{Preliminaries} 
A set of definitions must first be established before formally presenting the problem.

\smallskip\noindent\textbf{Definition 2.1 (Map)}. A map $\mathcal{M}$ represents the administrative boundaries of a finite and continuous geographic area of Earth, such as a city. Since $\mathcal{M}$ represents a relatively small region, the curvature of the Earth's surface within this area is negligible, allowing us to approximate $\mathcal{M}$ as a finite 2-dimensional Euclidean space $\mathbb{R}^2$.

\smallskip\noindent\textbf{Definition 2.2 (Trajectory)}. A trajectory represents the movement of a user or an object over time. It consists of a sequence of time-enabled spatiotemporal points denoted as $T = p_1p_2...p_n$, where each $p_i(\ell, t)$ represents a geolocation $\ell$ at time $t$.

\smallskip\noindent\textbf{Definition 2.3 (Partial trajectory)}. Given a trajectory $T$, an initial step $i$ and a length $l$, a partial trajectory is a subsequence $T_i^{l} = p_ip_{i+1}...p_{i+l-1}$ of $T$, where $p_i$ is the $i$'th spatiotemporal point in $T$.

\smallskip\noindent\textbf{Definition 2.4 (Trajectory history)}. The trajectory history of a specific length $T^{l}$ encompasses all the previously occurred partial trajectories of length $l$.

\smallskip\noindent\textbf{Definition 2.5 (Prediction horizon)}. The prediction horizon $k$ defines the number of future trajectory steps to be predicted.

\subsection{Problem Definition} 

We are now in a position to formally define the problem of estimating or forecasting the future path or trajectory of an object or entity based on its current state and historical data.

\smallskip\noindent \textbf{\textsc{Problem 1 (Trajectory prediction)}}.
Given a map $\mathcal{M}$, the corresponding trajectory history $T^{l}$ represented as a set of spatiotemporal point sequences, a partial trajectory $T^l_i = p_{i_1}p_{i_2}...p_{i_l}$ and a prediction horizon $k > 0$, the objective is to predict the next $k$ spatiotemporal points $p_{i_{l+1}},\dots, p_{i_{l+k}}$ of the partial trajectory $T^l_i$.

\smallskip\noindent Note that we currently only state the general trajectory prediction problem. Once we introduce the idea of higher-order mobility flow, we will revisit the trajectory prediction problem and formalize it in the context of predicting the next $k$ hexagons (see section \ref{sec:problem-revisited}).

\begin{table}[t!]
    \centering
    \begin{tabular}{c p{0.8\linewidth}}
        \toprule[1.2pt]
         \textbf{Symbol} & \textbf{Description}  \\
         \midrule
         $\mathcal{M}$ & The map of a geographic area  \\
         $\mathcal{B}$ & Set of hexagonal blocks forming a tessellation of $\mathcal{M}$ \\ 
         $T$ &  Trajectory sequence of spatiotemporal points\\
         $p_i$ & Spatiotemporal point in the trajectory sequence \\
         $T^l$ & Trajectory history of length $l$\\
         $k$ & Prediction horizon (\# trajectory steps to predict)\\ 
         $l$ & Input trajectory length\\
         \bottomrule[1.2pt]
    \end{tabular}
    \caption{Summary of key notations.}
    \label{tab:notations}
\end{table}

%% file: higher-order-trajectory-representations.tex
\section{Higher-order Mobility Flow Data}\label{sec:higher-order}

This section provides a rationale for working with higher-order mobility flow data. Additionally, we update the problem definition to accommodate the new data representation.

\smallskip\noindent\textbf{Rationale}. Working with raw trajectory datasets is challenging because GPS coordinates: ({\em i}) are sparse, and large amounts are needed to learn meaningful relationships, and ({\em ii}) are not very compatible (as input) with popular ML architectures, due to their continuous nature. We, therefore, propose to resort to a higher level of abstraction for representing trajectories. This transformation is done by first obtaining the routes/paths connecting raw trajectory data points through publicly available routing algorithms\footnote{Open Source Routing Machine (OSRM). \url{https://project-osrm.org/docs/v5.24.0/api/\#route-service}}. Then, the trajectory is represented as a sequence of the higher-order elements (hexagons) traversed by the route. We favor hexagons over other rectangular partitioning methods (such as Google S2\footnote{S2 Geometry Library. \url{http://s2geometry.io}} squares) because all six neighboring cells share identical properties, including equal distance to the cell's centroid and uniform border lengths. 
Moreover, hexagon-based tessellations offer several additional advantages: They provide a simpler and more symmetric definition of the nearest neighborhood, as each hexagon has six equidistant neighbors, eliminating the ambiguity inherent in rectangular grids that possess two types of neighbors (orthogonal and diagonal) with differing distances. Hexagons also approximate circles more closely than squares, resulting in a lower perimeter-to-area ratio (for instance, a unit-area hexagon has a perimeter of approximately 3.722 compared to 4 for a square), which minimizes edge effects. Furthermore, hexagonal grids exhibit greater isotropy, meaning that the relationship between grid-based and Euclidean distances varies less with direction, thereby reducing bias in spatial measurements and modeling dispersal and connectivity. 
This consistency makes hexagons more compatible with transformer architectures, as the transition from one token (hexagonal cell) to its neighbor would not be affected by any partitioning scheme. Additionally, mapping a point to its corresponding hexagonal token involves a constant-time operation through coordinate system conversions~\cite{h3geo}.

\smallskip\noindent\textbf{Definition 3.1 (Map Tessellation)}. Let $\mathcal{B}=\{b_1,b_2,...,b_n\}$ be a set of (regular) disjoint blocks that can fully tessellate the map $\mathcal{M}$, forming a regular tiling. Each block $b_k\in \mathcal{B}$ is assumed to be a polygon. In our study, we opt for {\em hexagons}. A hexagon-based map tessellation offers several advantages over a grid-based one (commonly known as a tile system) \cite{BIRCH2007347}. Note also that the tessellation can happen at different levels of resolution by defining different hexagon sizes; the smaller the hexagon size, the higher the resolution.

\begin{figure}[tb]
\includegraphics[width=0.87\textwidth]{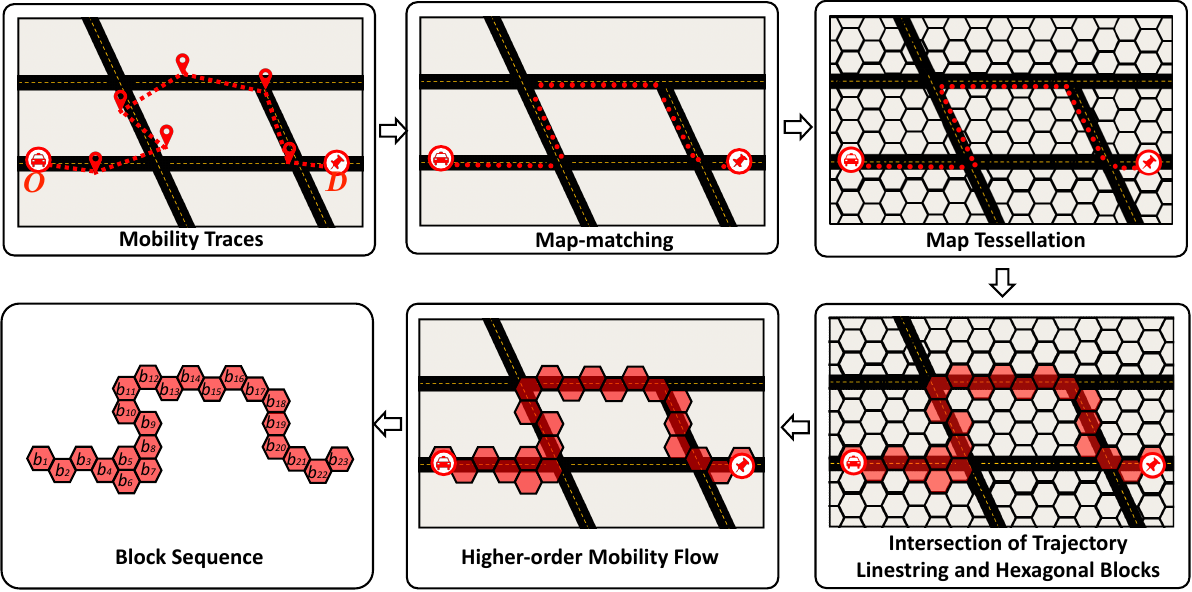}
 \caption{Construction of higher-order trajectory data.}
 \Description{A schematic diagram illustrating the process of constructing higher-order trajectory data. The figure depicts the transformation of raw trajectory inputs into enhanced representations.}
\label{fig:higher-order}
\end{figure}

\smallskip\noindent\textbf{Definition 3.2 (Higher-order Trajectory)}. Given a trajectory $T = p_1p_2...p_n$, and since every point $p_i$ resides within a unique block $b_i\in \mathcal{B}$ of a tessellated map $\mathcal{M}$, we can translate every trajectory as a sequence of blocks. By associating trajectory points with individual blocks, we imply that the predicted targets move step-wise, transitioning sequentially from one block to another. The outcome is a higher-order trajectory.

\smallskip\noindent The steps in transforming trajectory data points into sequences of hexagons are shown in Figure \ref{fig:higher-order}. Although the general pipeline is straightforward, it is not trivial and can be time-consuming. This is due to the involvement of specialized algorithms, such as {\em map-matching} and {\em computationally geometry} tasks.

\smallskip\noindent\textbf{Map-Matching.} The original trajectories are represented as a sequence of GPS-based data points. However, GPS data can be noisy and inaccurate, leading to deviations from actual roads. Map-matching aims to correct these inaccuracies and align the raw GPS points with the corresponding road network \cite{newson2009hmm}. While popular methods, such as \textsc{Ivmm} \cite{yuan2010ivmm}, exist for map-matching, one can use a routing machine like \textsc{Osrm} \cite{luxen2011osrm} to first find the shortest paths between consecutive points and then concatenate (in the same sequence) the shortest paths to form the map-matched trajectory.

\smallskip\noindent\textbf{Computational Geometry.} We utilize computational geometry methods to transform a map-matched trajectory into a sequence of hexagons. Every trajectory is modeled as a \texttt{linestring} shape type, and every hexagon as a 
\texttt{polygon} shape type. Then, their intersection can be computed using off-the-shelf methods of popular libraries. 
Recall that a map $\mathcal{M}$ can be tessellated using hexagons of different sizes, which defines the map's resolution. 
Upon data preparation, each trajectory in the dataset has been transformed into a sequence of hexagonal blocks (see Section~\ref{sec:dataset}). 

\subsection{Problem Definition (Revisited)}\label{sec:problem-revisited} 

Based on the introduction of higher-order trajectory representations, we now revisit the problem of trajectory prediction in the context of predicting $k$ future blocks (hexagons).

\smallskip\noindent \textbf{\textsc{Problem 2 (Higher-Order Trajectory Prediction)}}. Given a map $\mathcal{M}$, the corresponding trajectory history $T^{l}$ represented as a set of block sequences, a partial trajectory $T^l_i = b_{i_1}b_{i_2}...b_{i_l}$ (where $b_i$ is a block), and a prediction horizon $k > 0$, the objective is to predict the next $k$ blocks $b_{i_{l+1}},\dots,b_{i_{l+k}}$ of the partial trajectory $T^l_i$.

\smallskip\noindent Note that since higher-order mobility flow is defined at different levels of granularity, it offers a strategic trade-off between a trained model's accuracy and computational efficiency. The higher the resolution (i.e., the smaller the hexagons), the more refined the model's prediction, but the higher the training cost. This flexibility allows the model to be adapted to meet the needs of diverse applications.

%% file: trajectory-prediction-learning.tex
\section{Trajectory Prediction Learning}\label{sec:trajectory-prediction-learning}

In this section, we provide details of our \textsc{TrajLearn} model. In particular, we present ({\em i}) how our model leverages the Transformer architecture to capture intricate trajectory dependencies and facilitate accurate trajectory prediction, ({\em ii}) details of the model's training, and ({\em iii}) details of the beam search with constraints that allows to explore multiple possible future trajectory paths efficiently. Additionally, we discuss {\em model complexity} in Section~\ref{sec:complexity}. 

\begin{figure}[t!]
\includegraphics[width=0.99\textwidth]{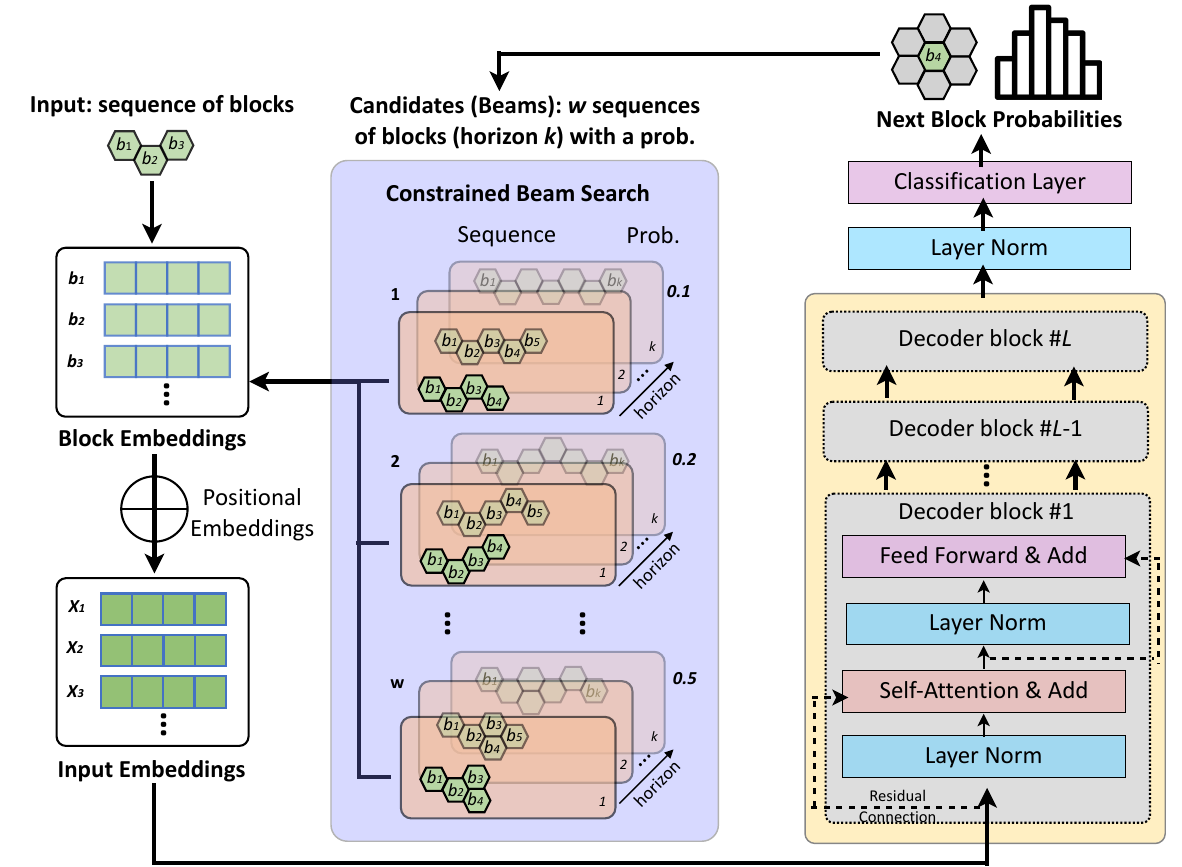}
 \caption{\textsc{TrajLearn} high-level architecture.}
 \Description{A schematic diagram illustrating the high-level architecture of the proposed framework.}
\label{fig:architecture}
\end{figure}

\subsection{Treating Trajectories as Statements}
To address our trajectory prediction task, we leverage the Transformer architecture \cite{vaswani2017attention} to capture underlying dependencies within trajectories. Though primarily designed for language tasks, Transformers are effective for our sequential trajectory data. The analogy can be outlined as follows: a \texttt{token} or \texttt{word} in language models corresponds to a \texttt{hexagon ID} (\texttt{hexagon}) in trajectory prediction. The number of \texttt{words} depends on vocabulary size; similarly, the number of \texttt{hexagons} depends on the map's tessellation. A \texttt{statement} is a sequence of \texttt{words}, just as a trajectory is a sequence of \texttt{hexagons}. Learning dependencies between \texttt{words} translates to learning dependencies between \texttt{hexagons}, enabling the Transformer to model complex sequential dependencies for trajectory prediction.
Figure \ref{fig:architecture} illustrates the training process framework.
In particular, our model is a $L$-layer decoder-only Transformer, each with $A$ causal self-attention heads and a $H$ dimensional state length. In contrast to the original Transformer architecture that used sinusoidal positional encodings, we used learned position embeddings. These embeddings are more flexible and capable of learning and adapting to complex patterns within the data. This way, the input to the Transformer is:
\vspace{-2pt}
\begin{equation}
h_0 = BW_e + W_p
\vspace{-2pt}
\end{equation}
where $B = (b_1, \dots, b_l)$ is the higher-order mobility flow, $W_e$ is the block embedding matrix, and $W_p$ is the position embedding matrix. 
Furthermore, unlike in the original Transformer, Layer normalization was moved to the input of each sub-block, and an additional Layer normalization was added after the final self-attention block. 
Formally, the computation of the hidden state at each Transformer layer $j\in [1, L]$ can be described as:
\begin{align}
  h'_j &= h_{j-1} + \texttt{Self-Attention}(\texttt{LayerNorm}(h_{j-1}))\\
      h_j &= h'_j + \texttt{FeedForward}(\texttt{LayerNorm}(h'_j))
\end{align}
where \texttt{LayerNorm$(\cdot)$}, \texttt{Self-Attention$(\cdot)$}, and \texttt{FeedForward}$(\cdot)$ denote layer normalization, the causal multihead self-attention operation, and the position-wise feed-forward network, respectively. For the \texttt{LayerNorm}, the module utilizes a modified version of L2 regularization proposed in \cite{loshchilov2017fixing} on all non-bias or gain weights. As an activation function, we opted for the Gaussian Error Linear Unit (\texttt{GELU}) \cite{hendrycks2016bridging}, which is chosen due to its performance in NLP tasks and its ability to alleviate the vanishing gradient problem, allowing for a more effective learning process. \texttt{GELU} is defined as:
\begin{equation}
\texttt{GELU}(x) = x \cdot P(X \leq x)
\end{equation}
where $X \sim N (0, 1)$ follows the standard normal distribution. In implementation, this is approximated by:
\begin{equation}
0.5x\left(1 + \tanh\left(\sqrt{\frac{2}{\pi}}\left(x + 0.044715x^3\right)\right)\right)
\end{equation}

In causal self-attention, every token is constrained to only attend to its left context. The attention mechanism can be formalized as:
\begin{equation}
\begin{gathered}
\texttt{Self-Attention}(E) = \texttt{softmax}\left(\frac{Q K^\top}{\sqrt{d_k}}  + \mathbf{M}
\right) V \\
\text{where} \quad \mathbf{M_{i,j}} = \begin{cases}
-\infty, & \text{$j > i$} \\
0, & \text{$j \leq i$}
\end{cases},
Q= E W_Q, \quad K = E W_K, \quad V = E W_V \\
\end{gathered}
\end{equation}

where $Q$, $K$ and $V$ are matrices representing the \texttt{queries}, \texttt{keys} and \texttt{values}, respectively, $W_Q$, $W_K$, and $W_V$ represent the respective learnable weight parameter matrices, and $d_k$ is the dimensionality of the \texttt{keys}. The matrix $\mathbf{M}$ is used to mask out future positions in the sequence, so the output of the self-attention layer for each position depends only on tokens to its left, ensuring causality in the attention mechanism. 
This mechanism allows the model to focus on different parts of the input sequence when generating the output.
The output of the last layer is fed into layer normalization, followed by a linear projection and a \texttt{softmax} activation that predicts the next block in the trajectory based on the probabilities of all possible next blocks:
\begin{equation}
    P(b_{l+1}|B) = \texttt{softmax}(\texttt{FeedForward}(\texttt{LayerNorm}(h_L)))
\end{equation}
Although we followed a decoder-only transformer architecture, our choice is motivated by the autoregressive nature of trajectory prediction, which requires the sequential generation of future spatial tokens conditioned solely on past trajectory data. A decoder-only transformer directly models the conditional distribution of each future hexagonal block given the preceding sequence, thereby simplifying both training and inference. This architecture also reduces computational overhead and enables seamless integration with our constrained beam search mechanism (\autoref{sec:beam}) to enforce spatial continuity between adjacent blocks. While encoder-decoder architectures may be beneficial for tasks involving more complex input-output mappings, our focus on the trajectory prediction task, renders the decoder-only approach particularly efficient. Nonetheless, our approach for trajectory prediction is flexible, and \textbf{other architectures used in large language models could be used}. Thus, any advancements in language models are applicable and can benefit our approach with minimal effort.

\subsection{Model Training}
\label{sec:model-training}

In the context of language models, the \texttt{<End of Sentence>} or \texttt{<EOS>} token serves as a special symbol or marker used to indicate the end of a sentence or sequence of words. 
In this work, we represent it with an \texttt{<End of Trajectory>} or \texttt{<EOT>} special token.
This step is vital as it enables the model to simulate real-world scenarios, where trajectories naturally conclude rather than continuing pointlessly. It also helps the model generate continuous trajectory paths where transitions mostly happen to adjacent hexagons. 
Once all trajectories have been represented as sequences of blocks, we transform them into {\em partial trajectories} that are required for training the prediction model. Specifically, we generate partial trajectories of length $l+k$, as well as all combinations of partial trajectories of length between $l$ and $l+k$. Note that $l\geq 1$ is a model parameter representing the input size, and $k\geq 1$ is the prediction horizon. 
Subsequently, the model is trained by providing the first $l$ blocks of partial trajectories as input and predicting the remaining ones. The process continues until an \texttt{<EOT>} token is predicted or until the trajectory has reached the prediction horizon.
 
\smallskip\noindent Our training procedure involves implementing a method known as {\em teacher forcing}. This technique is applied to stabilize the training process and accelerate convergence. During the training process, regardless of the model's current prediction of the next block, the correct block (i.e., the ground truth target block) is used to form the next time step's input. Figure \ref{fig:teacher-forcing} depicts this process. This approach provides a robust supervision signal and effectively allows the model to learn the latent dependencies and patterns of trajectories. We emphasize that the teacher forcing technique is used exclusively during the training phase and is not applied during inference.

\subsection{Beam Search with Constraints}
\label{sec:beam}

\begin{figure}
\centering
\begin{minipage}{.5\linewidth}
  \centering
    \captionsetup{width=.95\linewidth, font=small}
    \includegraphics[width=.92\linewidth, height=.95\linewidth]{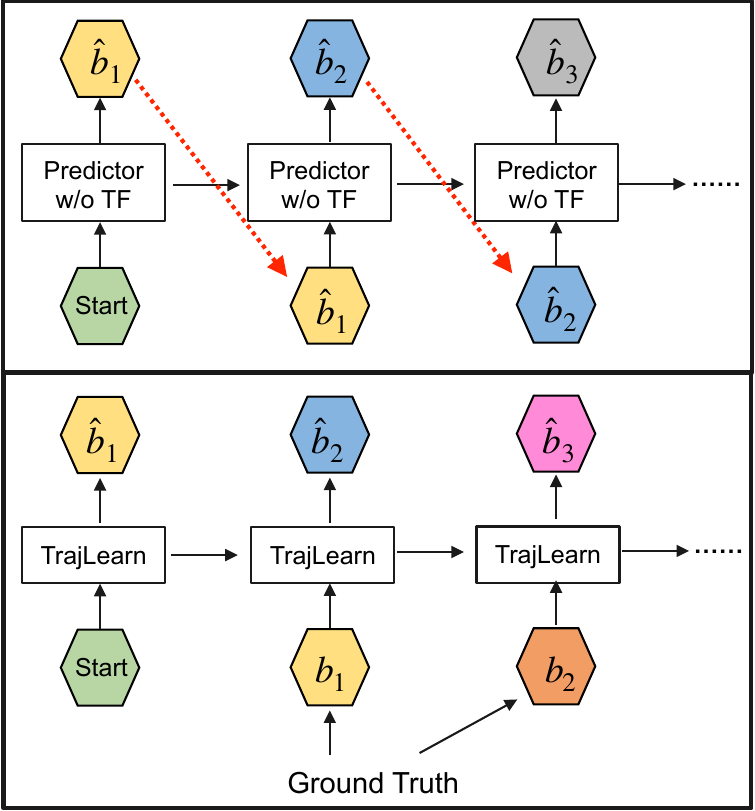}
    \caption{Train with (bottom) and w/o (top) teacher forcing.}
    \Description{Comparison of training with and without teacher forcing. The top diagram shows a model trained without teacher forcing, where predictions are fed back into the model. The bottom diagram shows training with teacher forcing, where ground truth is used at each step instead of predicted outputs.}
    \label{fig:teacher-forcing}
\end{minipage}%
\begin{minipage}{.5\linewidth}
  \centering
    \captionsetup{width=.95\linewidth, font=small}
    \includegraphics[width=.99\linewidth, height=.95\linewidth]{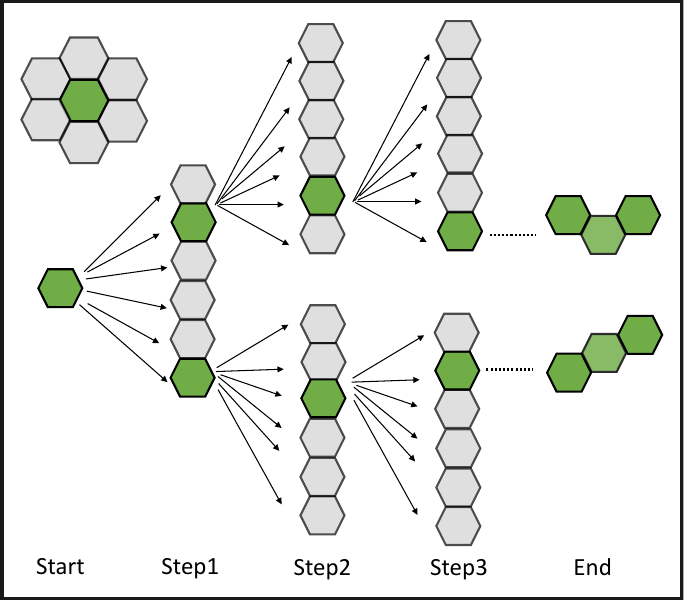}
    \caption{Beam search example where $w=2$ and $k=3$.}
    \Description{A beam search example with beam width 2 and sequence length 3. The figure illustrates multiple branching paths, showing how the top-k sequences are maintained at each step of the search process.}
    \label{fig:beam-search-example}
\end{minipage}

\vspace{-15pt}
\end{figure}

To optimize the trajectory prediction process and improve the performance, we incorporate {\em beam search with constraints}. 
Beam search is a heuristic search algorithm that explores the most promising trajectory paths.
It maintains a set of candidate sequences of blocks and generates new candidate sequences at each step by expanding the current best candidates. By guiding the search process and focusing on the most likely trajectory paths, beam search improves prediction performance and increases the quality of the predicted trajectories. Figure \ref{fig:teaser} shows an example of exploring two paths. Our model's beam search involves the following stages:
\begin{itemize}[leftmargin=*]
\item \textbf{Initialization.} The algorithm begins with the last visited block of the current trajectory as its initial state. It selects a set of candidate next blocks using the output probabilities provided by the model's classification layer.
\item \textbf{Beam Expansion.} Each candidate in the beam is expanded by one block, generating a new set of candidate blocks for the next step. The expansion process is guided by the spatial relationships between the blocks, allowing expansion only to geographically adjacent blocks based on the hexagon-based map tessellation,
The probabilities for each candidate in the beam are updated based on their cumulative probabilities as follows:
\vspace{-2pt}
\begin{equation}
P(b_{i_1}\dots b_{i_n}) = P(b_{i_1}\dots b_{i_{n-1}}) \times P(b_{i_n}|b_{i_1}\dots b_{i_{n-1}} )
\vspace{-2pt}
\end{equation}
\item \textbf{Beam Pruning.} After expansion, the beam is pruned to a certain beam width $w$, representing the most likely (or top) candidates for expansion based on their cumulative probabilities.
\item \textbf{Termination.} 
The algorithm continues iterating through the beam expansion and pruning stages until it either reaches the desired prediction horizon or all candidates encounter an \texttt{<EOT>}.
\end{itemize}
Figure \ref{fig:beam-search-example} depicts an illustrative example of the beam search with constraints with beam width $w=2$ and prediction horizon $k=3$.

\subsubsection{Ensuring the Validity of Predicted Trajectories}
To preserve {\em spatial continuity} in the predicted trajectories, we introduce constraints to the beam search algorithm. Formally, for a current block $b_i \in \mathcal{B}$, let $\Gamma_{\mathcal{M}}(b_i)$ denote the set of its adjacent blocks in $\mathcal{M}$. Then, at each step, the path can only expand from $b_i$ towards one of its adjacent blocks $b_j \in \Gamma_{\mathcal{M}}(b_i)$. By restricting the model to only consider adjacent blocks during the next block prediction task, we ensure {\em the validity} of the predicted trajectories, as they are always guided to follow spatially connected paths. This constraint further enhances the overall prediction {\em accuracy} by preventing the inclusion of non-adjacent blocks in the next block prediction task.

\subsection{Computational Complexity}
\label{sec:complexity}

\textsc{TrajLearn} relies on a Transformer architecture, the main operations of which include self-attention and feed-forward neural networks. During inference, we incorporate a beam search component. Below, we analyze the computational complexity of our model.

\smallskip\noindent\textbf{Self-Attention.}
The computational complexity of the self-attention mechanism is $O(n^2.d)$, where $n$ is the sequence length and $d$ is the dimension of the representation. Self-attention involves comparing each element in the sequence (like a word in a sentence) with every other element. This requires $n \times n$ comparisons or attention scores to be computed, leading to a quadratic complexity in terms of sequence length, and then, for each comparison, the model computes attention scores based on token representations, which are vectors of size $d$. This computation involves these vectors and hence introduces a factor of $d$ in the complexity.

\smallskip\noindent\textbf{Beam Search.} During the inference time, the beam search algorithm is utilized with a complexity of $O(w.\beta.k)$. Here, $w$ is the beam width, $k$ is the prediction horizon (depth of the search tree), and $\beta$ is the branching factor for each beam.
At each level of the tree, the algorithm examines $w$ nodes, and for each of these nodes, it considers up to $\beta$ child nodes. However, only the best $w$ among these $w.\beta$ nodes are retained for the next level. Therefore, the number of nodes evaluated at each level is proportional to $w.\beta$, and this process repeats for $k$ levels, which is our prediction horizon.

\smallskip\noindent\textbf{Hexagonal Tessellation.} Mapping a GPS coordinate to its corresponding hexagon is performed in constant time, i.e., $O(1)$, due to efficient coordinate-to-hexagon conversion methods. This ensures that the transformation from raw trajectory data to a hexagonal representation introduces minimal computational overhead. While the map-matching process, which aligns raw GPS data with the appropriate road network or spatial features, can be computationally intensive when processing large-scale data for training and testing, it is executed as part of an offline preprocessing pipeline. This separation guarantees that these heavy-lifting tasks do not impact the speed of the online inference process. Moreover, for real-time trajectory inference in practical applications, incremental map-matching techniques can be employed to update the trajectory representation dynamically as new data arrives, without significant overhead. Consequently, although the offline preprocessing stage may be resource-intensive, it is decoupled from real-time operations, ensuring that \textsc{TrajLearn} remains well-suited for online or real-time applications.

%% file: experiment.tex
\section{Experimental Evaluation}\label{sec:exp}

In this section, we offer a thorough experimental evaluation of our model, covering research questions, experimental setup, datasets, baseline methods, evaluation metrics, results, and insights. Our experiments aim to address the following questions, with the results presented in Subsection \ref{sec:experiment-results}. Additionally, we conduct an interpretability study in Subsection \ref{sec:interpretability} and map the predicted hexagons to GPS points in Subsection \ref{sec:hex2gps}.
\begin{enumerate}[label=\textbf{(Q\arabic*)}]
    \item \textbf{Model Accuracy Performance}. What is the accuracy performance of \textsc{TrajLearn} against sensible baselines?
    \item \textbf{Parameter Sensitivity Analysis}. How does the performance of our model vary with different input trajectory length $l$ and prediction horizon $k$?
    \item \textbf{Beam Search Analysis}. How is the performance of our model affected by varying values of the beam width $w$?
    \item \textbf{Map Resolution Analysis}. How does the performance of the model change with varying levels of map tessellation?
    \item \textbf{Ablation Study}. How does beam search with constraint and model architecture hyperparameters impact the model's performance?
\end{enumerate}

\subsection{Experimental Configuration}

\smallskip\noindent\textbf{Computational Environment}. 
We conducted experiments on a server equipped with an NVIDIA RTX A6000 graphics card and 320GB of memory. The model was developed in Python 3 and trained and deployed using the PyTorch 1.13 framework.

\smallskip\noindent\textbf{Map Tessellation and Resolutions}.
For tessellating a map, we utilized the \texttt{H3} geo-indexing system\footnote{H3 Geo-indexing Library. \url{https://h3geo.org/}}, which partitions the world into hexagonal cells of varying resolutions. We opted for \texttt{H3}'s resolutions 7, 8, and 9 for our experiments. 
Table \ref{tab:hexgonProp} reports on the hexagon's {\em edge length} (km) and surface area (km$^2$). 

\smallskip\noindent\textbf{Training Parameters}.
 We train our model using the AdamW optimizer, with an initial learning rate of $5\times 10^{-3}$, learning decay until reaching $5\times 10^{-7}$, batch size of $64$, and a dropout ratio of $0.1$.

\subsection{Datasets}
\label{sec:dataset}
In the experiments, we employ higher-order mobility flow data representations of three popular real-world trajectory datasets~\cite{10.1145/3589132.3625619}. We briefly describe the semantics of each dataset (prefixed by ``\textsc{Ho-}'' to indicate higher-order) and summarize their statistics in Table ~\ref{tab:data_stats}.
\begin{itemize}
    \item\textbf{\textsc{Ho-Porto}}~\cite{porto2015dataset}: This dataset consists of recorded mobility traces of taxis operating in Porto, Portugal. 

    \item \textbf{\textsc{Ho-Rome}} \cite{bracciale2014crawdadrome}: This dataset consists of recorded mobility traces of taxis operating in Rome, Italy.

    \item \textbf{\textsc{Ho-GeoLife}} \cite{zheng2008geolife, zheng2009geolife, zheng2010geolife}: This dataset consists of recorded mobility traces of individuals, covering a total distance of $\sim$1.2 million Km.
\end{itemize}
\smallskip\noindent After transforming each dataset into higher-order mobility flows, we split the trajectory dataset—ordered by the start times of the trajectories—into training, validation, and test sets using a 70\%, 10\%, and 20\% ratio, respectively. By partitioning the dataset into contiguous time intervals based on these start times, we eliminate the need to train and test over different random splits and report variance. This method ensures our evaluation reflects how the data naturally progresses over time, avoiding any randomness from dividing the data arbitrarily.

\begin{table*}[t!]
    \centering
      \begin{tabular}{ccc}
        \toprule[1.2pt]
         \textbf{\textsc{Resolution}} & \textbf{\textsc{Hex Edge Length (Km)}} & \textbf{\textsc{Hex Area (Km$^2$)}} \\
        \midrule
        \textbf{\textsc{Hex@7}}&	1.406& 5.161\\
        \textbf{\textsc{Hex@8}}&	0.531& 0.737\\
        \textbf{\textsc{Hex@9}}&	0.201& 0.105\\
      \bottomrule[1.2pt]
    \end{tabular}
    \caption[Statistics]{Properties of hexagons of different resolutions.}
    \label{tab:hexgonProp}
\end{table*}

\begin{table}[tb]
\centering
\begin{tabular}{lcccrrr}
\toprule[1.2pt]
\textsc{\textbf{Dataset}} & \textbf{\#\textsc{Entities}} & \textbf{\textsc{Observation Period}} & \textsc{\textbf{Res}} & \textsc{\textbf{\#Block}} & \textsc{\textbf{\#Trajectory}} & \textsc{\textbf{Avg. Length}} \\ 
\midrule
\multirow{3}{*}{\textsc{\textbf{Ho-Rome}}} & \multirow{3}{*}{315} & \multirow{3}{*}{02/01/14 -- 03/02/14}  & 7 & 172 & 5,678 & 72.75 \\
                               & & & 8 & 875 & 5,837 & 260.17 \\
                               & & & 9 & 4,231 & 5,854 & 689.75 \\ \hline
\multirow{3}{*}{\textsc{\textbf{Ho-Porto}}} & \multirow{3}{*}{442} & \multirow{3}{*}{07/01/13 -- 06/30/14} & 7 & 3,491 & 45,186 & 25.55 \\
                              & &  & 8 & 12,998 & 397,367 & 24.07 \\
                              & &  & 9 & 45,633 & 1,151,544 & 35.33 \\ \hline
\multirow{3}{*}{\textsc{\textbf{Ho-GeoLife}}} & \multirow{3}{*}{52} & \multirow{3}{*}{04/01/07 -- 10/31/11} & 7 & 1,878 & 1,556 & 117.58 \\
                              &  &   & 8 & 6,360 & 1,830 & 219.28 \\
                               &  &  & 9 & 21,270 & 1,964 & 525.72 \\ 
\bottomrule[1.2pt]
\end{tabular}
\caption{Statistics of the processed datasets.}
\label{tab:data_stats}
\end{table}

\subsubsection{Training Data}
 \textsc{TrajLearn} is trained using the Higher-order mobility datasets at various resolutions (7, 8, and 9), as discussed in Section~\ref{sec:higher-order}. To ensure data quality, we excluded trajectories consisting of fewer than 15 hexagonal blocks. This exclusion was necessary because the experiment setup requires a minimum of $l = 10$ historical data points to forecast subsequent $k = 5$ points. Consequently, the number of trajectories in our training dataset differs from the original datasets. Table~\ref{tab:data_stats} presents the statistics for the training datasets across these resolutions.

\subsection{Baseline Methods}
The datasets discussed in this research are commonly used in trajectory prediction research, but comparing results across studies has become challenging due to several reasons: ({\em i}) some prior research has introduced additional metadata, which was not initially present in the original dataset, such as POI data~\cite{8741193} or weather data~\cite{9304734}. These additions can favor certain models over others, irrespective of their architecture or training methods; ({\em ii}) different researchers employ various preprocessing pipelines, leading to significant differences in the generated data. This can involve excluding trajectories with GPS measurement errors~\cite{10.5555/3056172.3056180, 9304734} or outliers~\cite{9312472, 9304734}, which often represent the most challenging trajectories to predict.
Therefore, to assess the performance of our model, we have selected prime models from existing literature that do not depend on additional meta information to serve as baselines. Each model represents a different approach to trajectory prediction.

\smallskip\noindent\textbf{\textsc{MC}} \cite{10.1145/2181196.2181199}. A Markov Chain (MC) is a commonly used model for sequence prediction. It considers each location as a state and makes predictions based on a transition matrix between these states.

\smallskip\noindent\textbf{\textsc{LSTM}} \cite{650093}. Long Short-Term Memory (LSTM) is a type of Recurrent Neural Network (RNN) specifically designed to capture long-term sequential dependencies, which are essential for mobility prediction tasks. In our experiments, we explored a range of hyperparameters, including: \texttt{embedding size} (64, 100, 200, 300, 500), \texttt{hidden size} (64, 128, 200), \texttt{number of layers} (1 or 2), and \texttt{dropout rate} of 0.2 for the middle layer in a 2-layer LSTM. We report the best results.

\smallskip\noindent\textbf{\textsc{LSTM-ATTN}} \cite{luong2015effective}. LSTM with attention is a variant of LSTM that integrates an attention mechanism that allows the model to focus on specific parts of the sequence when making predictions. We applied the same parameter search space that we used for the LSTM.

\smallskip\noindent\textbf{\textsc{GRU}} \cite{GRU}. Gated Recurrent Units (GRU) is a variant of RNN designed to address the vanishing gradient problem of RNNs and can be applied to sequence prediction tasks. We used the same parameters range for this model as for the LSTM.

\smallskip\noindent\textbf{\textsc{DeepMove}} \cite{feng2018deepmove}. \textsc{DeepMove} is a state-of-the-art method combining a multi-modal recurrent network and a historical attention mechanism to capture both spatial and temporal dependencies.

\smallskip\noindent\textbf{\textsc{Flashback++}} \cite{10.1145/3616541}. \textsc{Flashback++} is a follow-up work by the authors of \textsc{Flashback} \cite{yang2020location}. It is a system that relies on RNN and uses sparse semantic trajectory modeling to predict the next location by looking for similar trajectories in terms of temporal characteristics. A grid search was conducted on the hidden dimension values $\{10, 32, 64, 128, 256\}$, and the optimal value was selected based on the results.

\smallskip\noindent\textbf{Baselines Implementation.} For the baselines, we implemented the \textsc{MC}, \textsc{LSTM}, \textsc{LSTM-ATTN}, and \textsc{GRU} models ourselves. For the \textsc{DeepMove} model, we used the implementation provided by \cite{libcity}\footnote{\url{https://github.com/LibCity/Bigscity-LibCity}}, and for \textsc{Flashback++} we relied on the implementation by \cite{10.1145/3616541}\footnote{\url{https://github.com/Pursue1221/FlashbackPlusPlus}}, both with adjustments to support hexagonal-based spatial data in our experiments. Computational complexity, memory complexity, and the number of trainable parameters comparison of baseline models and \textsc{TrajLearn} is reported in \autoref{tab:complexity}.

\begin{table*}[t!]
\centering
\resizebox{\textwidth}{!}{%
\begin{tabular}{c|l|cccc|cccc|cccc}
\toprule[1.5pt]
& & \multicolumn{4}{c|}{\textsc{\textbf{Resolution 7}}} & \multicolumn{4}{c|}{\textsc{\textbf{Resolution 8}}} & \multicolumn{4}{c}{\textsc{\textbf{Resolution 9}}}\\
\textsc{Dataset}&\textsc{Model} & \textsc{Acc@1} & \textsc{Acc@3} & \textsc{Acc@5} & \textsc{BLEU} &  \textsc{Acc@1} & \textsc{Acc@3} & \textsc{Acc@5} & \textsc{BLEU} & \ \textsc{Acc@1} & \textsc{Acc@3} & \textsc{Acc@5} & \textsc{BLEU} \\
\midrule
\parbox[t]{2mm}{\multirow{7}{*}{\rotatebox[origin=c]{90}{\textbf{\textsc{Ho-Porto}}}}}
&\textbf{\textsc{MC}}           & 0.2232 & 0.2460 & 0.2483 & 0.2443 & 0.2131 & 0.2390 & 0.2426 & 0.2359 & 0.2239 & 0.2511 & 0.2561 & 0.2490 \\
&\textbf{\textsc{LSTM}}         & 0.4360 & 0.5070 & 0.5325 & 0.4836 & 0.3268 & 0.4435 & \underline{0.4888} & 0.3778 & 0.3744 & 0.5168 & 0.5626 & 0.4132 \\
&\textbf{\textsc{LSTM-ATTN}}   & 0.4383 & 0.5112 & 0.5387 & 0.4891 & 0.3233 & 0.4377 & 0.4781 & 0.3719 & 0.3675 & 0.5086 & 0.5556 & 0.4075\\
&\textbf{\textsc{GRU}}        & 0.3140 & 0.3741 & 0.4089 & 0.3581 & 0.2010 & 0.2771 & 0.3067 & 0.2411 & 0.2236 & 0.3074 & 0.3364 & 0.2588 \\
&\textbf{\textsc{DeepMove*}} & 0.0802 & 0.2038 & 0.3450 & 0.1516 & 0.1272 & 0.1994 & 0.2482 & 0.1725 & 0.2463 & 0.3139 & 0.3497 & 0.2858 \\
 &\textbf{\textsc{Flashback++*}}     & \underline{0.4439} & \underline{0.5015} & \underline{0.5254} & \underline{0.4929} & \underline{0.3320} & \underline{0.4599} & 0.4867 & \underline{0.4066} & \underline{0.3993} & \underline{0.5316} & \underline{0.5809} & \underline{0.4314} \\
&\textbf{\textsc{TrajLearn (ours)}}      & \textbf{0.4507} & \textbf{0.5285}& \textbf{0.5648}& \textbf{0.5108}& \textbf{0.4244}& \textbf{0.5638}& \textbf{0.6138}& \textbf{0.4860}& \textbf{0.4785}& \textbf{0.6555}& \textbf{0.7111}& \textbf{0.5255} \\
\cmidrule{1-14}
\rowcolor{lightgray!10}&\textbf{Improvement (\%)}          & \textbf{{1.53}} & \textbf{{5.38}} & \textbf{{7.50}} & \textbf{{3.63}} & \textbf{{27.84}} & \textbf{{22.59}} & \textbf{{26.11}} & \textbf{{19.53}} & \textbf{{19.81}} & \textbf{{23.30}} & \textbf{{22.42}} & \textbf{{21.80}} \\
\midrule
\parbox[t]{2mm}{\multirow{7}{*}{\rotatebox[origin=c]{90}{\textbf{\textsc{Ho-Rome}}}}}
&\textbf{\textsc{MC}}           & 0.0440 & 0.0591 & 0.0643 & 0.0685 & 0.1335 & 0.1482 & 0.1512 & 0.1504 & 0.1459 & 0.1699 & 0.1726 & 0.1686 \\
&\textbf{\textsc{LSTM}}         & 0.2284 & 0.2919 & 0.3195 & 0.2566 & 0.3191 & 0.4152 & 0.4536 & 0.349 & 0.3664 & 0.5020 & 0.5527 & 0.3977 \\
&\textbf{\textsc{LSTM-ATTN}}   & 0.2264 & 0.2892 & 0.3170 & 0.2550 & 0.3164 & 0.4133 & 0.4508 & 0.3462 & 0.3663 & 0.5016 & 0.5522 & 0.3972 \\
&\textbf{\textsc{GRU}}        & 0.2132 & 0.2656 & 0.2934 & 0.2392 & 0.2244 & 0.2806 & 0.3064 & 0.2480 & 0.1636 & 0.2420 & 0.2801 & 0.1932 \\
&\textbf{\textsc{DeepMove}}    & \underline{0.2644} & \underline{0.3594} & \underline{0.3940} & \underline{0.2966} &  \underline{0.3529} & 0.4140 & 0.4367 & \underline{0.3815} & OOM & OOM & OOM & OOM \\
 &\textbf{\textsc{Flashback++}}     & 0.2448 & 0.3086 & 0.3386 & 0.2658 & 0.3364 & \underline{0.4292} & \underline{0.4666} & 0.3634 & \underline{0.3860} & \underline{0.5269} & \underline{0.5821} & \underline{0.4225} \\
&\textbf{\textsc{TrajLearn (ours)}}      & \textbf{0.2924} & \textbf{0.3700}& \textbf{0.4046}& \textbf{0.3279}& \textbf{0.3953}& \textbf{0.5149}& \textbf{0.5631}& \textbf{0.4317}& \textbf{0.4515}& \textbf{0.6085}& \textbf{0.6704}& \textbf{0.4886} \\
\cmidrule{1-14}
\rowcolor{lightgray!10}&\textbf{Improvement (\%)}          & \textbf{{10.60}} & \textbf{{2.95}} & \textbf{{2.69}} & \textbf{{10.56}} & \textbf{{17.50}} & \textbf{{19.95}} & \textbf{{20.70}} & \textbf{{18.80}} & \textbf{{16.96}} & \textbf{{15.51}} & \textbf{{15.15}} & \textbf{{15.63}} \\
\midrule
\parbox[t]{2mm}{\multirow{7}{*}{\rotatebox[origin=c]{90}{\textbf{\textsc{Ho-GeoLife}}}}}
&\textbf{\textsc{MC}}           & 0.1045 & 0.1083 & 0.1093 & 0.1113 & 0.0743 & 0.0848 & 0.0858 & 0.0864 & 0.0665 & 0.0882 & 0.0899 & 0.0857 \\
&\textbf{\textsc{LSTM}}         & 0.3837 & 0.4710 & 0.5009 & 0.3996 & 0.3632 & 0.4325 & 0.4631 & 0.3787 & 0.3909 & 0.4898 & 0.5203 & 0.4166 \\
&\textbf{\textsc{LSTM-ATTN}}   & 0.4208 & 0.4840 & 0.5109 & 0.4376 & 0.4081 & 0.4742 & 0.5048 & 0.4271 & 0.3892 & 0.4809 & 0.5136 & 0.4142 \\
&\textbf{\textsc{GRU}}        & 0.3051 & 0.3544 & 0.3995 & 0.3183 & 0.2187 & 0.3135 & 0.3656 & 0.2391 & 0.1070 & 0.1811 & 0.2437 & 0.1308 \\
&\textbf{\textsc{DeepMove}}    & \underline{0.4212} & \underline{0.5928} & \underline{0.6679} & \underline{0.4765} & 0.3598 &  \underline{0.5136} & \underline{0.6255} & 0.3778 & OOM & OOM & OOM & OOM \\
 &\textbf{\textsc{Flashback++}}     & 0.3907 & 0.4755 & 0.5072 & 0.4072 & \underline{0.3911} & 0.4420 & 0.4885 & \underline{0.3995} & \underline{0.4144} & \underline{0.5154} & \underline{0.5301} & \underline{0.4311} \\
&\textbf{\textsc{TrajLearn (ours)}}      & \textbf{0.6008} & \textbf{0.6683}& \textbf{0.7028}& \textbf{0.6235}& \textbf{0.5303}& \textbf{0.6082}& \textbf{0.6427}& \textbf{0.5565}& \textbf{0.4266}& \textbf{0.5247}& \textbf{0.5589}& \textbf{0.4545} \\
\cmidrule{1-14}
\rowcolor{lightgray!10}&\textbf{Improvement (\%)}          & \textbf{{42.60}} & \textbf{{12.75}} & \textbf{{5.23}} & \textbf{{30.82}} & \textbf{{35.60}} & \textbf{{37.63}} & \textbf{{31.56}} & \textbf{{39.30}} & \textbf{{2.94}} & \textbf{{1.80}} & \textbf{{8.44}} & \textbf{{5.43}} \\
\bottomrule[1.5pt]
\end{tabular}%
}

\caption{\textsc{TrajLearn} accuracy performance against five baselines, for varying evaluation metric and resolution, over three benchmark datasets. We fix input length $l=10$ \& prediction horizon $k=5$. The bold/underlined numbers indicate the best/second best method, respectively. Improvement (\%) reports the relative improvement of our model over the strongest baseline. (*Experiments with \textsc{DeepMove} on \textsc{Ho-Porto} and \textsc{Flashback++} on \{\textsc{Ho-Porto},  resolution 9\} were conducted on 30,000 randomly sampled trajectories due to their limited efficiency and scalability on large datasets. )
}
\label{tab:main}
\end{table*}

\begin{table}[ht]
    \centering
    \begin{tabular}{p{0.22\linewidth} p{0.23\linewidth} p{0.23\linewidth} p{0.22\linewidth}}
        \toprule[1.2pt]
         \textbf{Model} & \textbf{Time Complexity} & \textbf{Memory Complexity} & \textbf{\# Parameters} \\
         \midrule        
         \textsc{MC} & \(O(1)\) & \(O(1)\) & $\approx 10K$ \\
         \textsc{LSTM} & \(O(T \cdot H^2)\) & \(O(T \cdot H)\) & $\approx 6.1M$ \\
         \textsc{LSTM-ATTN} & \(O(T \cdot H^2 + T^2 \cdot H)\)& \(O(T^2 \cdot H)\) & $\approx 6.1M$\\
         \textsc{GRU} & \(O(T \cdot H^2)\) & \(O(T \cdot H)\) & $\approx 5.1M$ \\
         \textsc{DeepMove} & \(O(T \cdot H^2 + T^2 \cdot H)\) & \(O(T^2 \cdot H)\) & $\approx 8.4M$ \\
         \textsc{Flashback++} & \(O(T \cdot H^2)\) & \(O(T \cdot H)\) & $\approx 6.5M$ \\
         \textsc{TrajLearn} & \(O(T \cdot H^2 + T^2 \cdot H)\) & \(O(T^2 \cdot H)\) & $\approx 7.3M$ \\
         \bottomrule[1.2pt]
    \end{tabular}
    \caption{Computational complexity, memory complexity, and number of trainable parameters of baseline models and \textsc{TrajLearn} for a single inference. Here, \(T\) is the sequence length, and \(H\) is the hidden size. The trainable parameter count is reported for models trained on \textsc{Ho-GeoLife}, res=7.}
    \label{tab:complexity}
\end{table}

\subsection{Evaluation Metrics}

To evaluate the performance of our model against the baselines, we consider and adopt the following well-established metrics for evaluating sequence prediction tasks of trained language models. 

\smallskip\noindent\textbf{\textsc{Accuracy@N}} $[\uparrow]$. This metric assesses how often the correct sequence appears within the top-$N$ ranked predictions made by the model.
Given a set of sequences $P$ in the test dataset, with a sequence denoted as $s$, the actual label of $s$ as $\text{true}(s)$, and the set of top $N$ predictions for $s$ as $\text{Top}_N(s)$, it is defined as:
\begin{equation}
    \textsc{Accuracy@N} = \frac{|\{ s \in P \mid \text{true}(s) \in \text{Top}_N(s) \}|}{|P|}
\end{equation}
This metric evaluates a model's ability to include the true label in its top $N$ predictions, summarizing performance at various precision levels. We report \textsc{Accuracy@1}, \textsc{Accuracy@3}, and \textsc{Accuracy@5}.

\smallskip\noindent\textbf{\textsc{BLEU score}} $[\uparrow]$. This is a standard metric for sequence prediction tasks in NLP, which measures the quality of predicted sequences by comparing them with the ground truth sequences. In our context, we had to adapt the BLEU score to assess predicted trajectories. 
We do so by quantifying the similarity between the predicted trajectories (sequences of blocks) and the actual trajectories by comparing the overlap of $n$-grams (or $n$-blocks) -- contiguous sequences of $n$ blocks along the trajectories -- between them, incorporating a brevity penalty for overly short predicted trajectories.
For a given trajectory, an $n$-block consists of a specific sequence of blocks (e.g., a turn at an intersection followed by a straight path and another turn), analogous to a combination of words forming a meaningful sentence. 
Formally, the BLEU score in our context is given by:
\[ \text{BLEU} = BP \cdot \exp\left(\sum_{n=1}^{T} w_n \log p_n\right) \]
where $p_n$ is the ratio of number of $n$-blocks matches to the total number of $n$-blocks in the predicted trajectories, $w_n$ are weights that sum to 1 ($\sum_{n=1}^{T} w_n = 1$), and $T$ is the maximum order of $n$-block considered. The brevity penalty $BP$ is defined as follows:
\begin{equation}
BP = \begin{cases}
1 & \text{if } c > r \\
e^{(1 - r/c)} & \text{if } c \leq r
\end{cases}
\end{equation}
where $c, r$ are the lengths of the predicted and ground truth sequences, respectively.
In our experiments, we adhere to NLP best practices and consider up to 4-blocks ($T=4$) and uniform weights.

\subsection{Results and Discussion}
\label{sec:experiment-results}

\smallskip\noindent\textbf{(Q1) Model Accuracy Performance.} \label{section:q1} 
We compare the performance of \textsc{TrajLearn} against the baselines employing different metrics and varying resolutions over three real-world trajectory datasets. We fix the input trajectory length ($l=10$) and prediction horizon ($k=5$). 
Table \ref{tab:main} shows the numerical results, where for each metric, the \textbf{bold} and \underline{underlined} numbers correspond to the best and second-best performing model, respectively.
A few key observations can be made: (i) \textsc{TrajLearn} demonstrates a remarkable performance by consistently securing one of the top two spots and, in all instances outperforming all competitors by a large margin (see \% improvement), 
(ii) \textsc{TrajLearn}'s accuracy is improving as the $N$ of the \textsc{Accuracy@N} is increasing, where $N$ represents the number of top predictions considered. This is due to the {\em teacher forcing} technique involved in the training stage (see Section~\ref{sec:model-training}), which provides supervision and corrects the prediction for any subsequent step, effectively allowing \textsc{TrajLearn} to learn. 

\smallskip\noindent Note that \textsc{DeepMove} encounters out-of-memory (OOM) issues with large datasets, specifically \textsc{Ho-Porto} at all resolutions, \textsc{Ho-Rome} at resolution 9, and \textsc{Ho-GeoLife} at resolution 9. To address this, we conducted experiments on randomly sampled trajectories ($\sim$30,000) for each dataset, similar to the practice in \textsc{Flashback++} \cite{10.1145/3616541}. However, for \textsc{Ho-Rome} and \textsc{Ho-GeoLife} at resolution 9, the sample sizes were too small to achieve satisfactory performance.
The main reason for this is the method's approach to training the prediction model, which heavily depends on sparsely available POI check-in datasets. Upon aligning their methodology with our continuous path approach (hexagonal traversal), the volume of ``check-ins'' surges substantially, leading to OOM errors. Despite these OOM errors hindering some experiments, their impact on the overall conclusions is minimal, as the remaining results demonstrate our approach's generalizability and practicality. Our model preprocesses and trains on data batches, eliminating the need to load the entire dataset into memory. This reduces the memory footprint and enhances efficiency when processing large datasets. 

\begin{figure}[t!]
    \centering
    \includegraphics[width=0.94\textwidth]{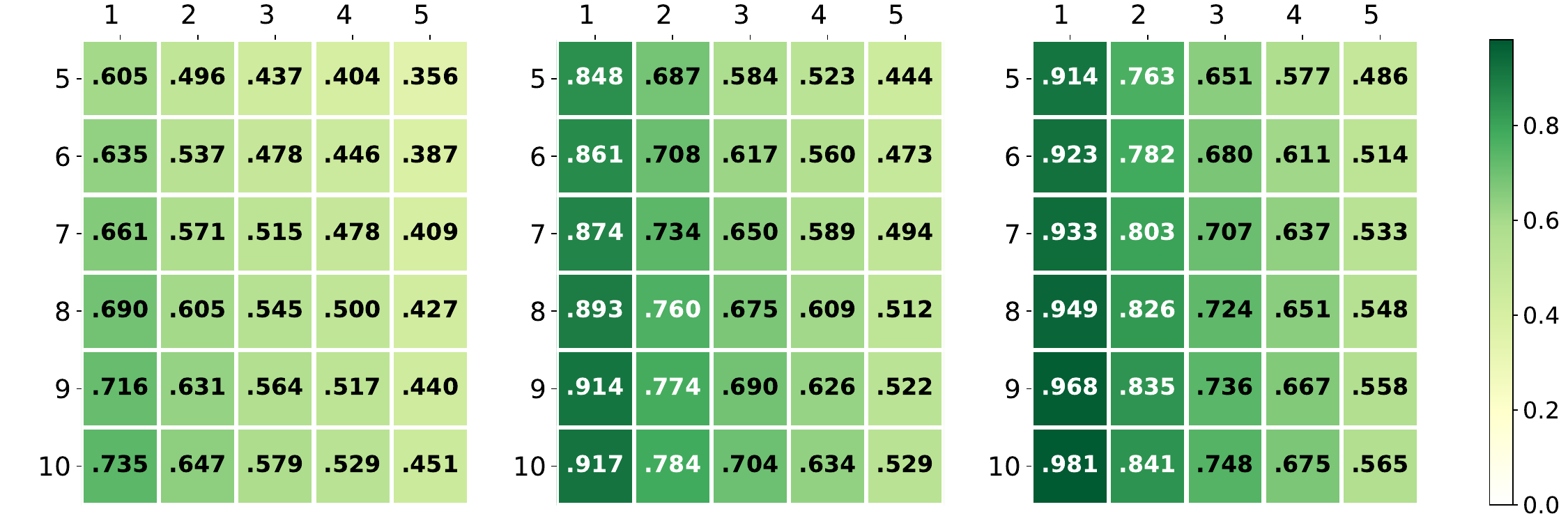}
    \begin{minipage}[t]{.3\linewidth}
        \centering
        \vspace{-2em}
        \caption*{\hspace{-2.25em}(a) Accuracy@1}
    \end{minipage}%
    \begin{minipage}[t]{.3\linewidth}
        \centering
        \vspace{-2em}
        \caption*{\hspace{-2.25em}(b) Accuracy@3}
    \end{minipage}
    \begin{minipage}[t]{.3\linewidth}
        \centering
        \vspace{-2em}
        \caption*{\hspace{-2.25em}(c) Accuracy@5}
    \end{minipage}
    \caption{\textsc{TrajLearn} accuracy for varying prediction horizon $k$ (horizontal) \& input length $l$ (vertical) on \textsc{Ho-Porto}, res=7.}
    \Description{A heatmap visualization divided into three panels displaying the performance of the TrajLearn model on the Ho-Porto dataset (resolution 7). Panel (a) shows Accuracy@1, panel (b) shows Accuracy@3, and panel (c) shows Accuracy@5. The horizontal axis represents the prediction horizon (k) while the vertical axis represents the input length (l). Color intensity in each panel indicates the accuracy achieved by the model.}
    \label{fig:sensitivity}
\end{figure}

\smallskip\noindent\textbf{(Q2) Parameter Sensitivity Analysis}. In this experiment, we investigate the influence of varying input trajectory length ($5\leq l \leq 10$) and prediction horizon ($1\leq k \leq 5$) on the accuracy of \textsc{TrajLearn} at different precision levels. 
Figure \ref{fig:sensitivity} presents the results for \textsc{Ho-Porto} with resolution 7.
A few key observations can be made: (i) the longer the trajectory history $l$, the higher the prediction accuracy; (ii) the shorter the prediction horizon $k$, the higher the accuracy prediction. These trends are logical, as predicting a far horizon with limited information as input becomes increasingly challenging. 
Similar to \textbf{(Q1)}, the accuracy is improving as the $N$ of the \textsc{Accuracy@N} is increasing.

\begin{figure}[t!]
    \centering
    \begin{subfigure}[b]{0.54\textwidth}
         \centering
     \hspace*{-1.5em}\includegraphics[width=\textwidth]{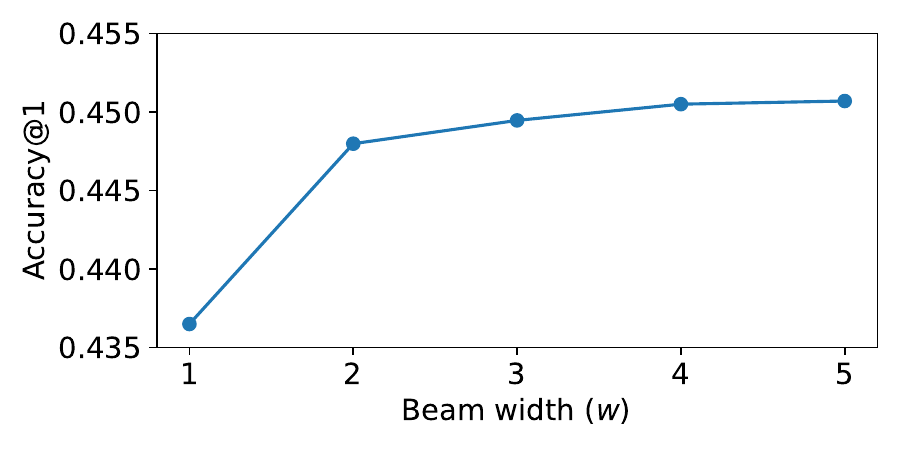}
    \end{subfigure}
    \hspace*{0.02\textwidth}
    \begin{subfigure}[b]{0.27\textwidth}
         \centering
     \hspace*{-1.5em}\includegraphics[width=\textwidth]{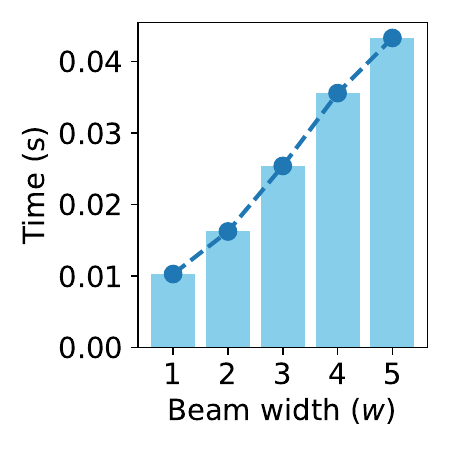}
    \end{subfigure}
    \caption{Impact of beam width $w$ on \textsc{TrajLearn}'s accuracy (left) and inference time (right) on \textsc{Ho-Porto}, res=7 and batch size of 64}.
    \Description{A two-panel figure illustrating the effect of beam width on TrajLearn's performance using the Ho-Porto dataset. The left panel displays a plot showing how the model's accuracy changes with different beam widths, while the right panel depicts the corresponding inference time. The plots are generated with a resolution of 7 and a batch size of 64, highlighting the trade-off between accuracy and computational cost as the beam width increases.}
    \label{fig:beam-search}
\end{figure}

\smallskip\noindent\textbf{(Q3) Beam Search Analysis}. In this experiment, we investigate the tradeoff between {\em the accuracy performance} of \textsc{TrajLearn} and its {\em running cost during inference}, as a result of varying values of the beam width $w$ ranging from 1 to 5. Figure \ref{fig:beam-search} shows the results for \textsc{Accuracy@1} (left) and inference time averaged per batch (right) on \textsc{Ho-Porto} with resolution 7. The findings suggest that augmenting the beam width typically boosts the model's performance as anticipated. However, this enhancement becomes less apparent as the number of beams increases, implying diminishing returns.
The running cost during inference increases linearly with the beam width, as expected. Consequently, we set the beam width to $w=5$ (and not larger) for the rest of our experiments.

\begin{figure}[t!]
    \centering
    \begin{subfigure}[b]{0.48\textwidth}
         \centering
          \hspace*{5pt}\includegraphics[height=7.5em]{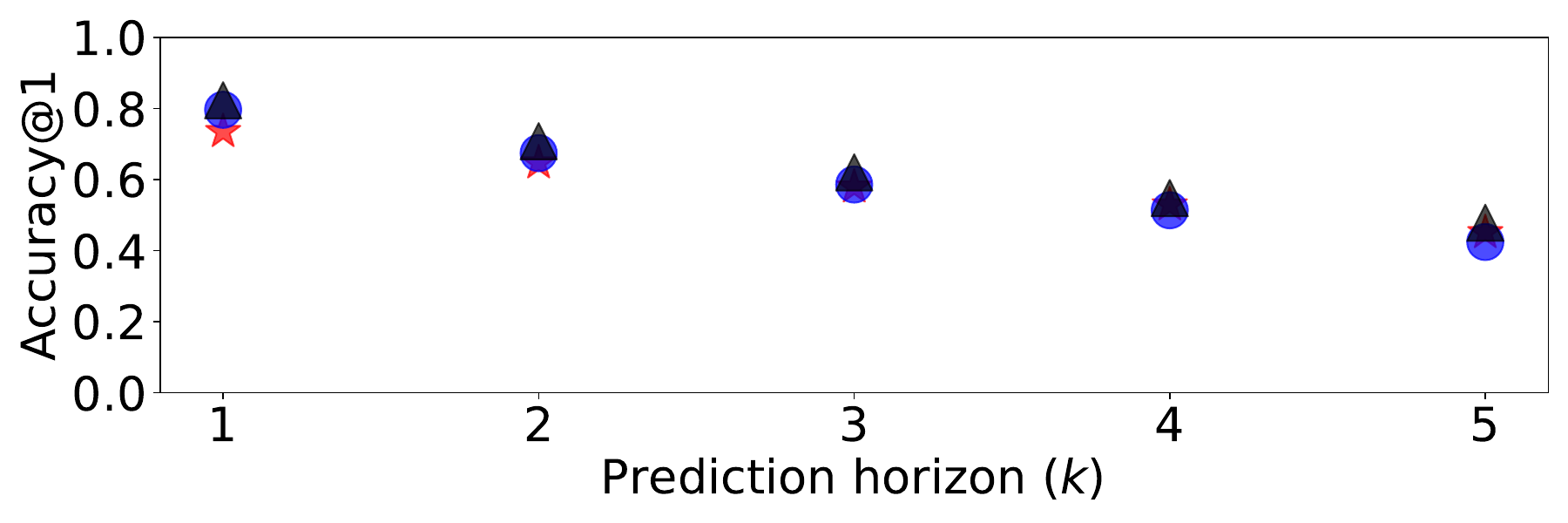}
     \end{subfigure}
     \begin{subfigure}[b]{0.48\textwidth}
         \centering
          \hspace*{-5pt}\includegraphics[height=7.5em]{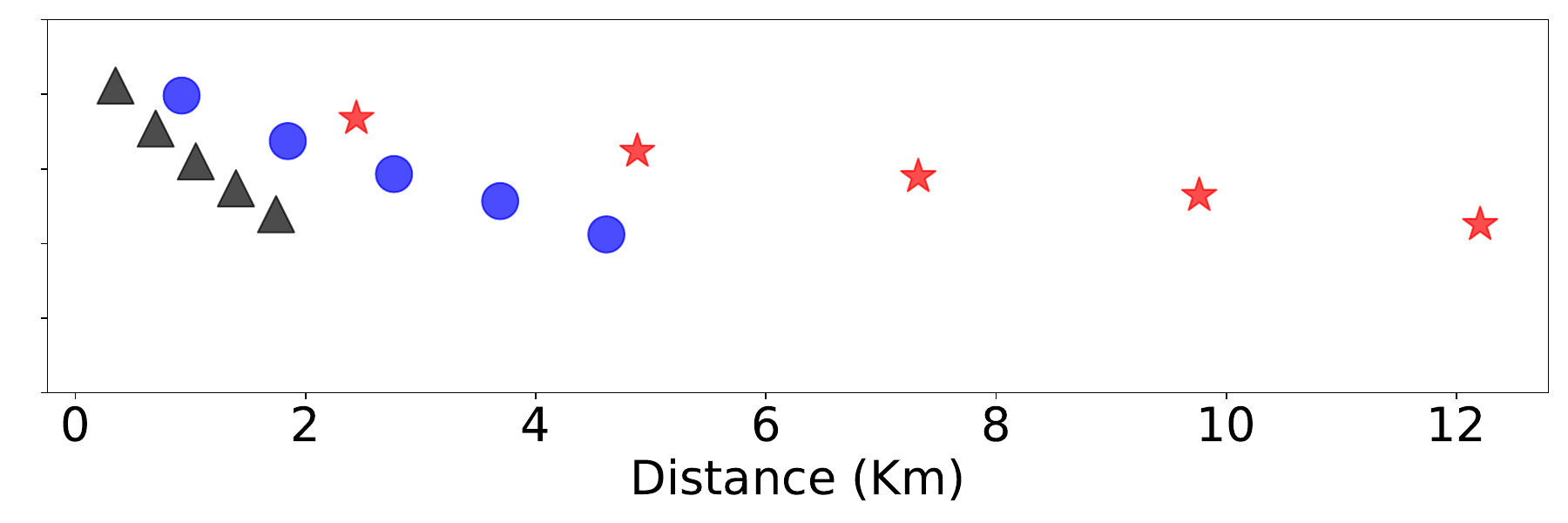}
     \end{subfigure}
     \begin{subfigure}[b]{0.48\textwidth}
         \centering
          \hspace*{5pt}\includegraphics[height=7.5em]{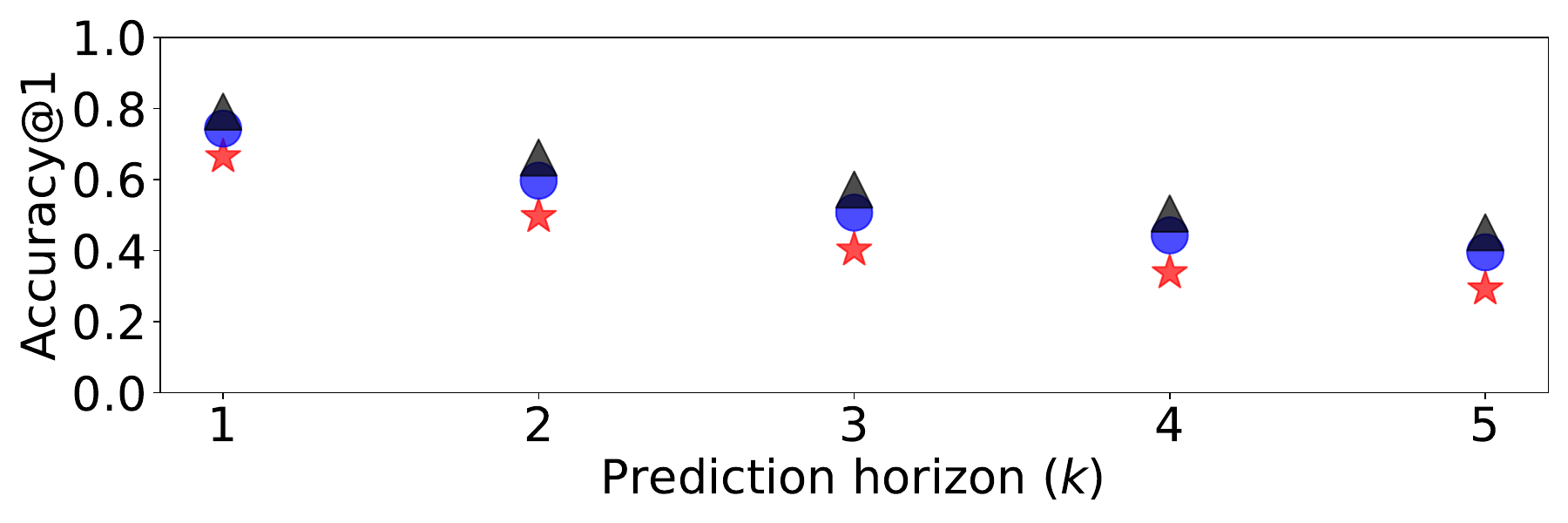}
     \end{subfigure}
     \begin{subfigure}[b]{0.48\textwidth}
         \centering
          \hspace*{-5pt}\includegraphics[height=7.5em]{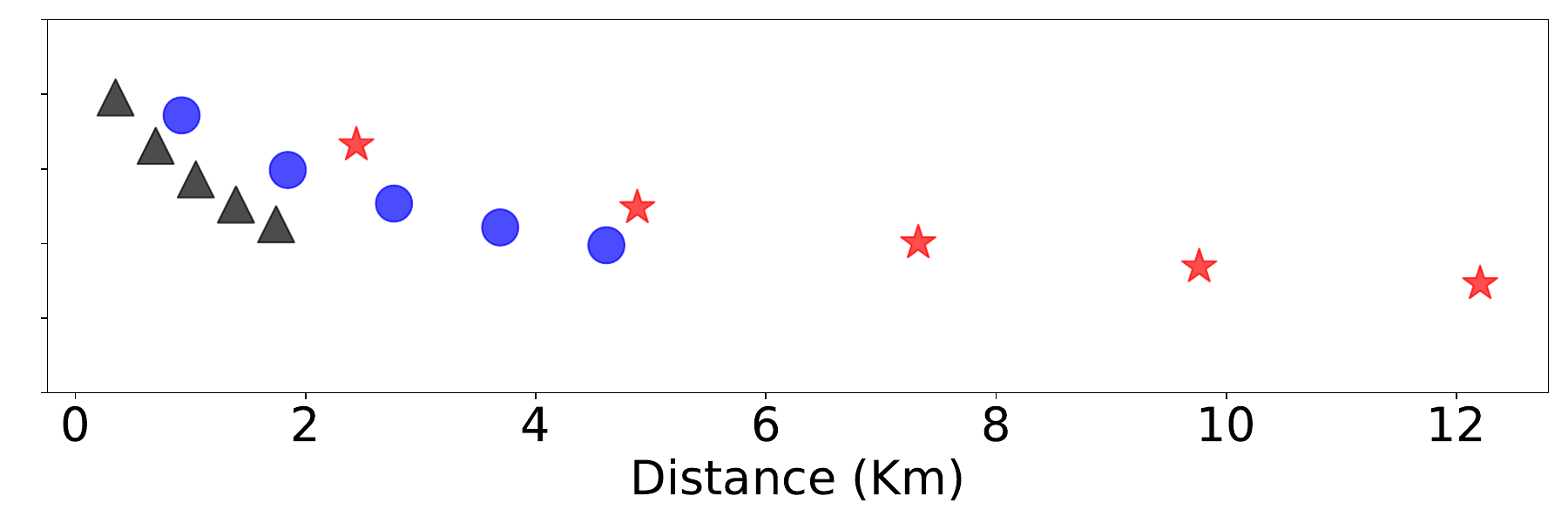}
     \end{subfigure}
     \begin{subfigure}[b]{0.48\textwidth}
         \centering
          \hspace*{5pt}\includegraphics[height=7.5em]{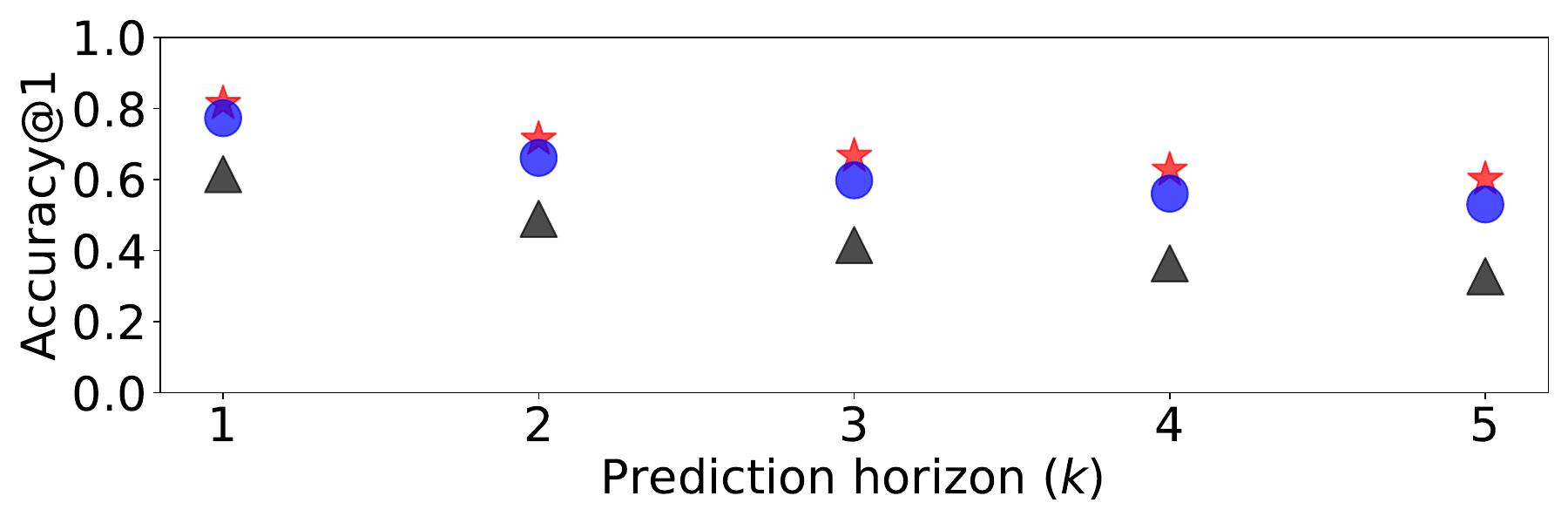}
     \end{subfigure}
     \begin{subfigure}[b]{0.48\textwidth}
         \centering
          \hspace*{-5pt}\includegraphics[height=7.5em]{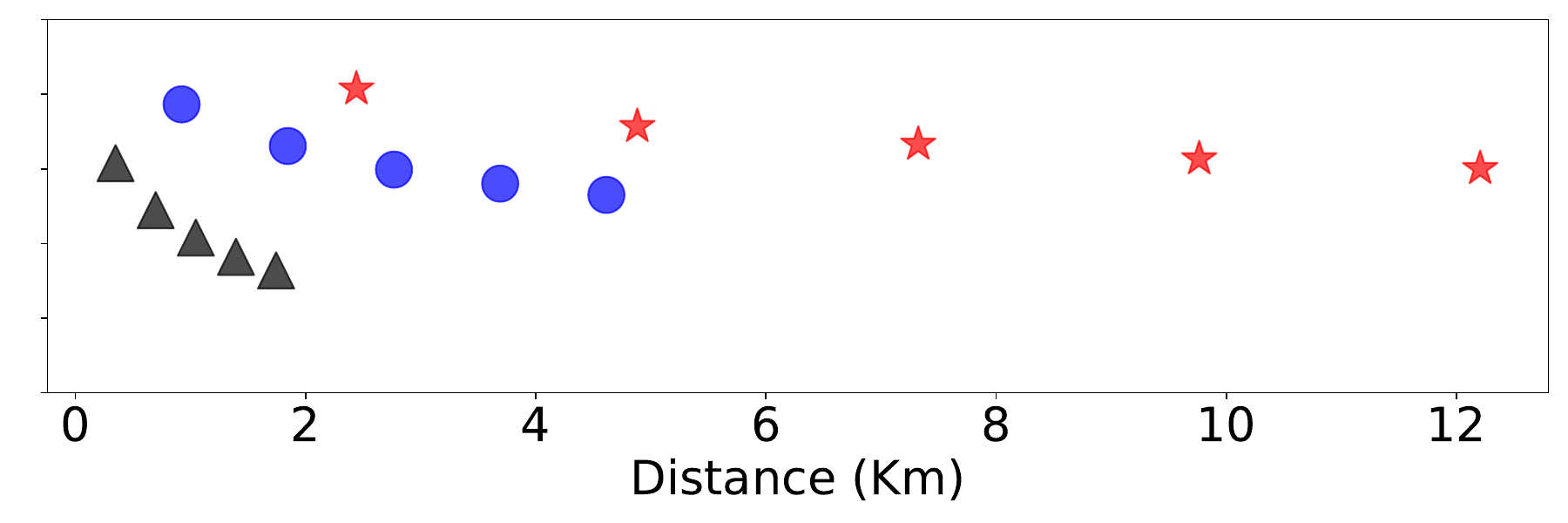}
     \end{subfigure}
    \caption{\textsc{TrajLearn} accuracy for varying resolutions (\textcolor{red}{7: $\star$}, \textcolor{blue}{8: $\bullet$}, \textcolor{black}{9: $\blacktriangle$}) on \textsc{Ho-Porto} (top), \textsc{Ho-Rome} (middle), \textsc{Ho-GeoLife} (bottom). We report \textsc{Accuracy@1} as a factor of the prediction horizon $k$ (left) and the actual distance traveled (right).}
    \Description{Six-panel figure arranged in two columns and three rows. Each row corresponds to a different dataset: the top row shows results for Ho-Porto, the middle row for Ho-Rome, and the bottom row for Ho-GeoLife. In each row, the left panel plots Accuracy@1 as a function of the prediction horizon, while the right panel plots Accuracy@1 as a function of the actual distance traveled. Different markers and colors denote the resolution levels: red star for resolution 7, blue bullet for resolution 8, and black triangle for resolution 9.}
    \label{fig:accuracy-resolution}
\end{figure}

\smallskip\noindent\textbf{(Q4) Map Resolution Analysis}. In this experiment, we investigate the impact of a map's resolution on \textsc{TrajLearn}'s performance. Recall that a lower (higher) resolution means larger (smaller) hexagons. Depending on how the data is collected (e.g., vehicles or individuals) and the specific domain application, the different resolutions offer a trade-off between computational efficiency and accuracy. For illustration purposes, Figure~\ref{fig:accuracy-resolution} presents the results for varying resolutions on the \textsc{Ho-Porto} dataset. A few observations can be made: (i) the smaller the resolution, the slower the rate with which the accuracy decreases as the prediction horizon increases; (ii) the smaller the resolution, the slower the rate with which the accuracy decreases, as the distance traveled increases.

\begin{figure}
    \centering
    \begin{subfigure}[b]{0.33\textwidth}
         \centering
         \captionsetup{width=.95\linewidth, font=small}
          \includegraphics[width=\textwidth]{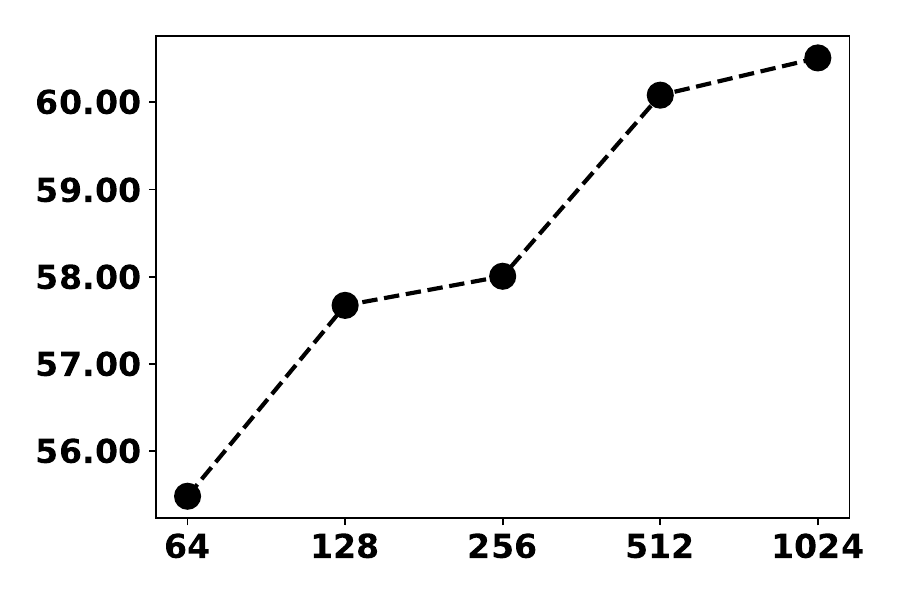}
          \includegraphics[width=\textwidth]{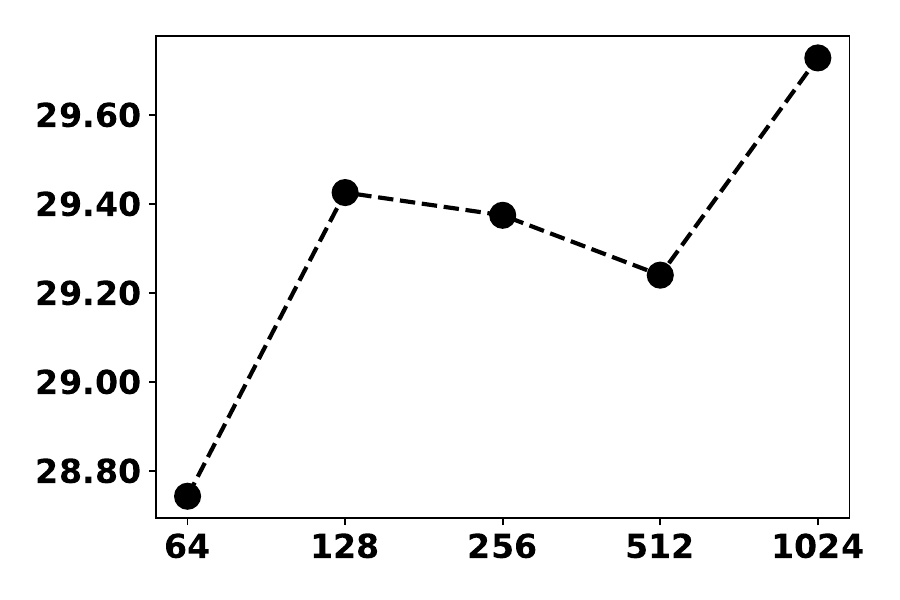}
          \includegraphics[width=\textwidth]{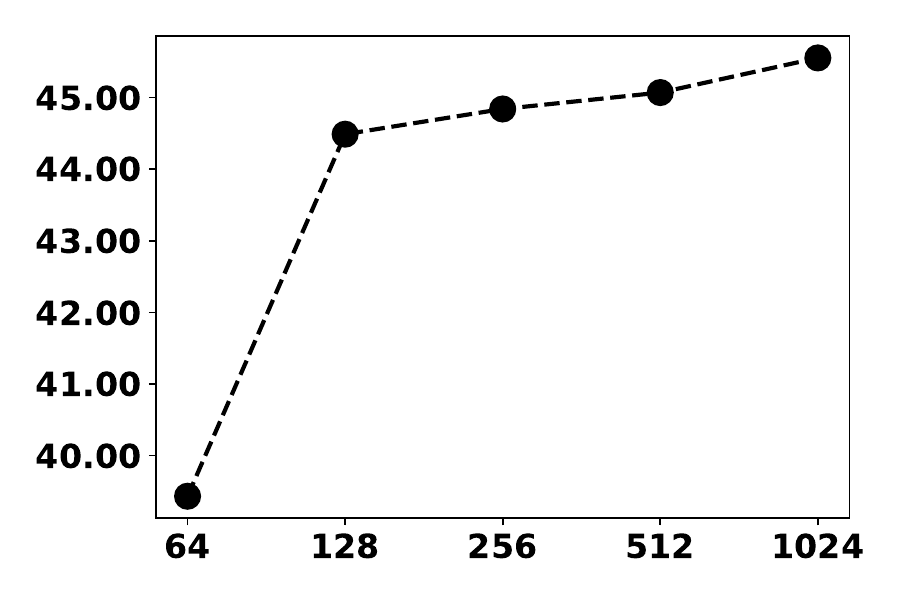}
     \end{subfigure}
     \begin{subfigure}[b]{0.33\textwidth}
         \centering
          \includegraphics[width=\textwidth]{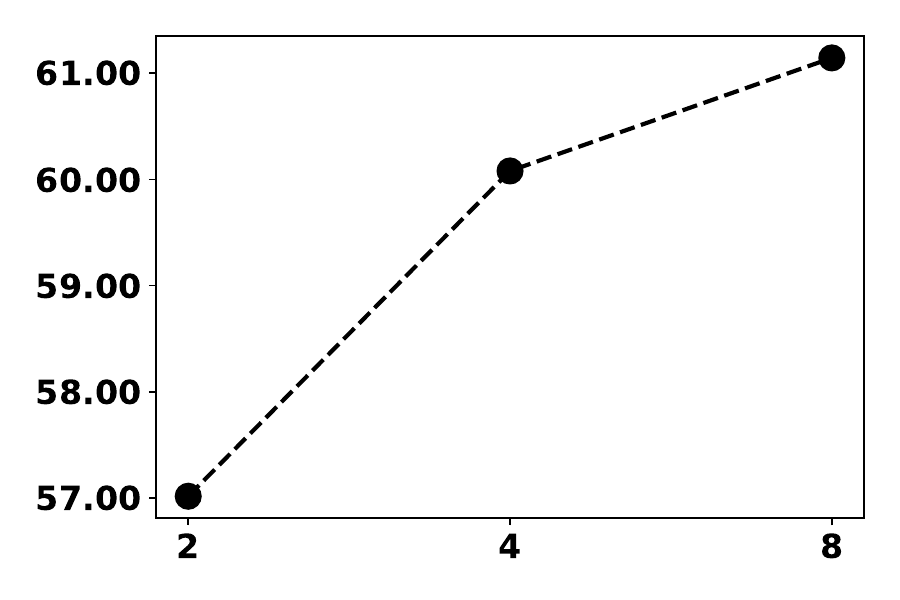}
          \includegraphics[width=\textwidth]{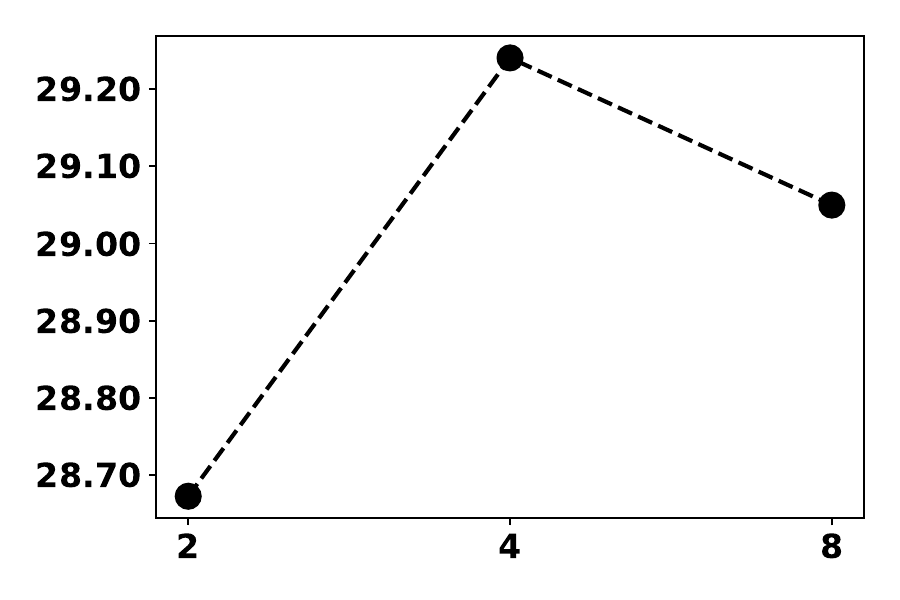}
          \includegraphics[width=\textwidth]{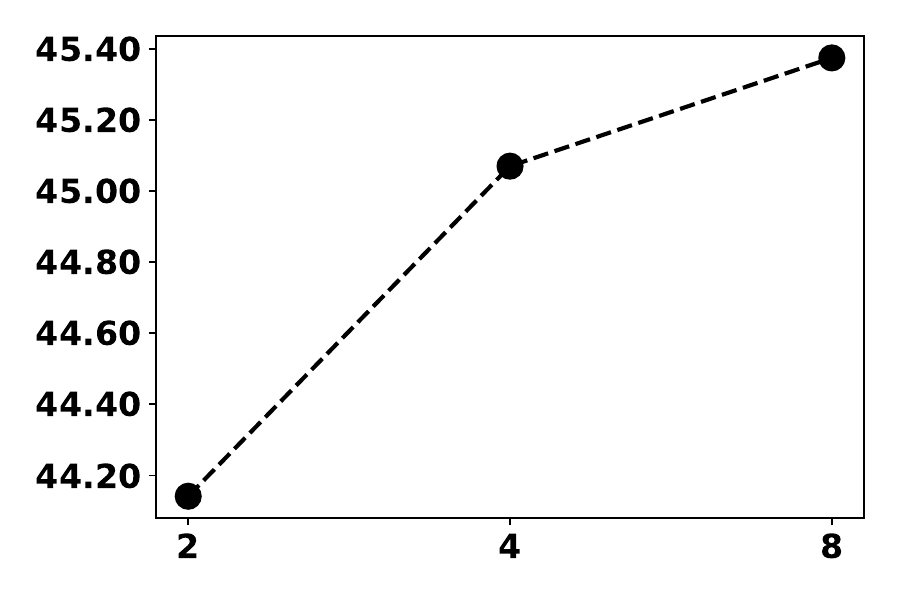}
     \end{subfigure}
     \begin{subfigure}[b]{0.33\textwidth}
         \centering
          \includegraphics[width=\textwidth]{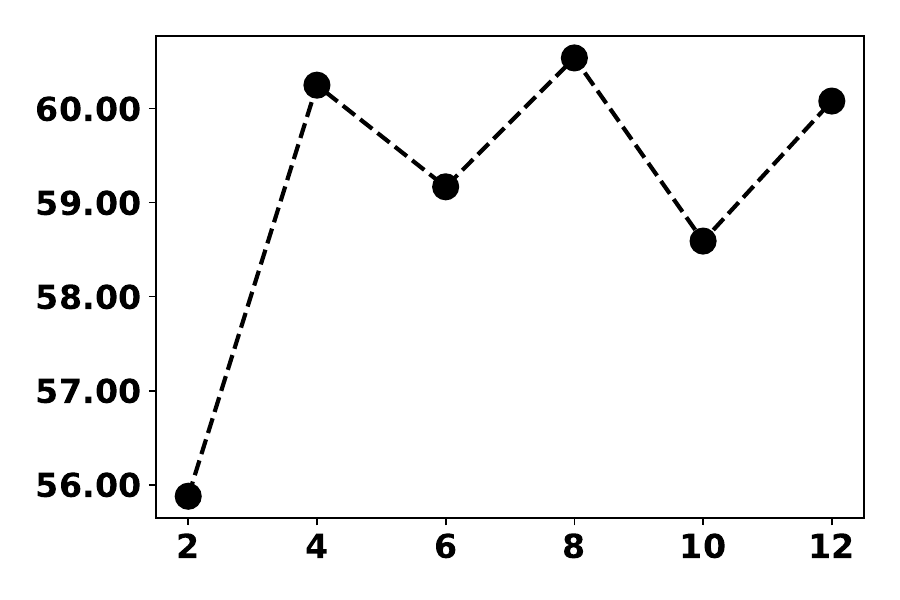}
          \includegraphics[width=\textwidth]{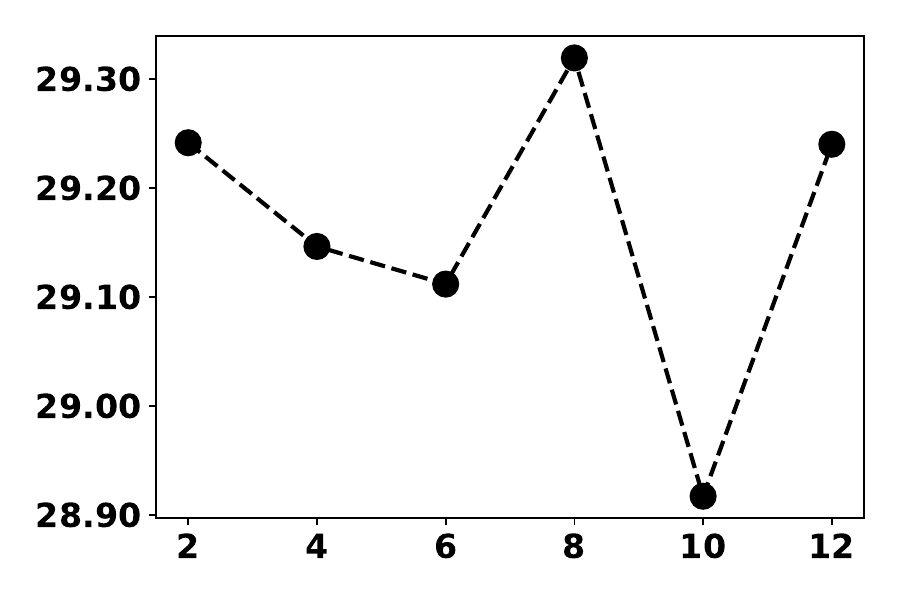}
          \includegraphics[width=\textwidth]{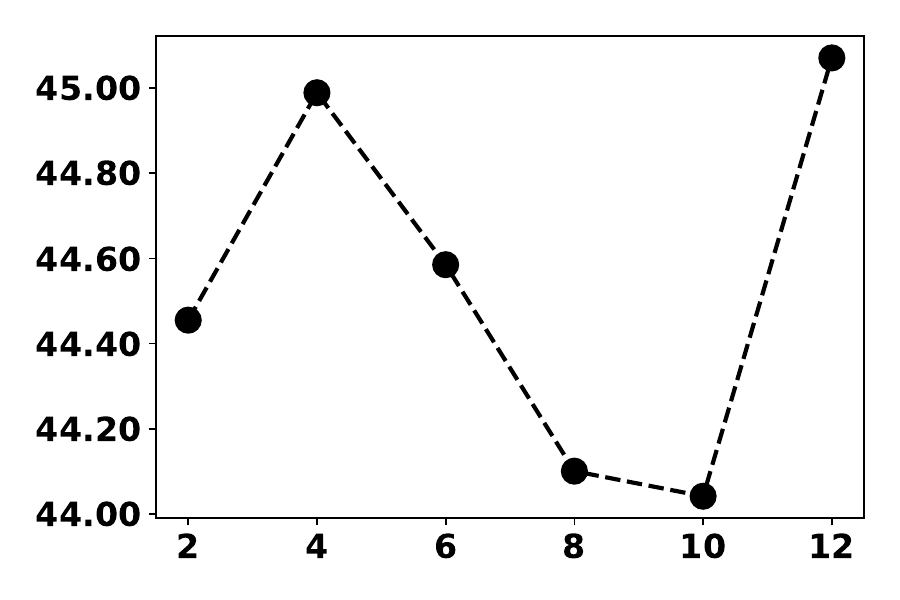}
     \end{subfigure}
    \caption{\textsc{Accuracy@1} over datasets \textsc{Ho-GeoLife} (top), \textsc{Ho-Rome} (middle), and \textsc{Ho-Porto} (bottom) with resolution 7 for varying embedding vector size (left), number of attention heads (middle), and number of Transformer layers (right).}
    \Description{A three-column figure composed of nine panels arranged in three rows. The top row shows results for the Ho-GeoLife dataset, the middle row for Ho-Rome, and the bottom row for Ho-Porto, all at resolution 7. The left column displays plots of Accuracy@1 with varying embedding vector sizes, the middle column shows plots for different numbers of attention heads, and the right column presents plots for varying numbers of Transformer layers.}
    \label{fig:ablation}
\end{figure}

\smallskip\noindent\textbf{(Q5) Ablation Study}. In this experiment, we investigate the impact of certain components of \textsc{TrajLearn}'s neural architecture. We selectively remove the beam search module and report the {\em accuracy performance change}. Table~\ref{tab:ablation} shows the results, which indicate that the accuracy drops when removing the beam search. In addition, we perform a 
hyperparameter analysis to gauge the impact of various parameters on accuracy: {\em embedding dimension}, {\em number of decoder layers}, and {\em number of attention heads}. These tests were carried out using the \textsc{Ho-GeoLife} dataset at resolution 7. Figure~\ref{fig:ablation} presents the results. From our observations, \textsc{TrajLearn}'s performance benefits from an increase in the number of layers, as it allows for a more comprehensive processing of dependencies. Likewise, enhancing the number of attention heads results in a better outcome. Additionally, increasing the embedding dimension enhances performance up to a certain threshold, likely due to the constrained input length.

\begin{table}[t!]
    \centering
      \begin{tabular}{cccc}
        \toprule[1.2pt]
         \textbf{\textsc{Dataset}} & \textbf{\textsc{Accuracy@1}} & \textbf{\textsc{Accuracy@1} W/O Beam}&\textbf{Change (\%)} \\
        \midrule
        \textbf{\textsc{Ho-Porto@7}}& 0.4507 &	0.4365 & -3.15\\ 
        \textbf{\textsc{Ho-Porto@8}}& 0.4244 &	0.4064 & -4.24\\ 
        \textbf{\textsc{Ho-Porto@9}}& 0.4785 &	0.4677 & -2.26\\
      \bottomrule[1.2pt]
    \end{tabular}
    \caption[Statistics]{Impact of removing the beam search on accuracy.}
    \label{tab:ablation}
\end{table}

\smallskip\noindent\textbf{Discussion}. Beyond addressing the five experimental evaluation questions, we now delve deeper into the underlying reasons for \textsc{TrajLearn}’s performance improvements. Unlike conventional autoregressive approaches, such as RNN-based models (e.g., LSTM and GRU), which often suffer from vanishing gradients and limited long-range context, \textsc{TrajLearn} leverages a transformer-based architecture that models the joint distribution of future trajectories, thereby effectively capturing complex higher-order mobility flows. The inherent self-attention mechanism of transformers enables the model to learn intricate spatial-temporal dependencies over extended sequences, seamlessly integrating cues from both local (recent) contexts and long-range dependencies that are critical for accurate prediction, thus yielding more informed and precise outcomes. Furthermore, the model’s generative formulation facilitates the exploration of multiple plausible future paths in a coherent and probabilistic manner, enhancing its ability to account for the inherent uncertainty and multimodality in movement data.
Moreover, although hexagonal tessellation is common across all evaluated models, its superior ability to capture subtle spatial dependencies—thanks to its uniform neighborhood structure—provides a richer and more consistent representation of the movement space. This enhanced spatial representation is particularly well-exploited by a powerful model like the transformer, which can discern nuanced relationships both between adjacent spatial units and across larger regions.
Additionally, the employment of teacher forcing during training guides the model with ground-truth sequences, thereby promoting more robust learning of complex spatial-temporal patterns.
Finally, the integration of a constrained beam search mechanism during inference enables \textsc{TrajLearn} to explore multiple plausible future paths in a coherent and probabilistic manner, ensuring that predicted trajectories adhere to spatial continuity constraints. This approach effectively accounts for the inherent uncertainty and multimodality in movement data, ultimately generating realistic and feasible paths.
Collectively, these design choices empower \textsc{TrajLearn} to handle complex higher-order mobility flows, leading to performance improvements of up to approximately 40\% over baseline methods, as demonstrated in our experiments.

\subsection{Interpretability Study}
\label{sec:interpretability}

\begin{figure*}[tb]
    \includegraphics[width=0.95\textwidth]{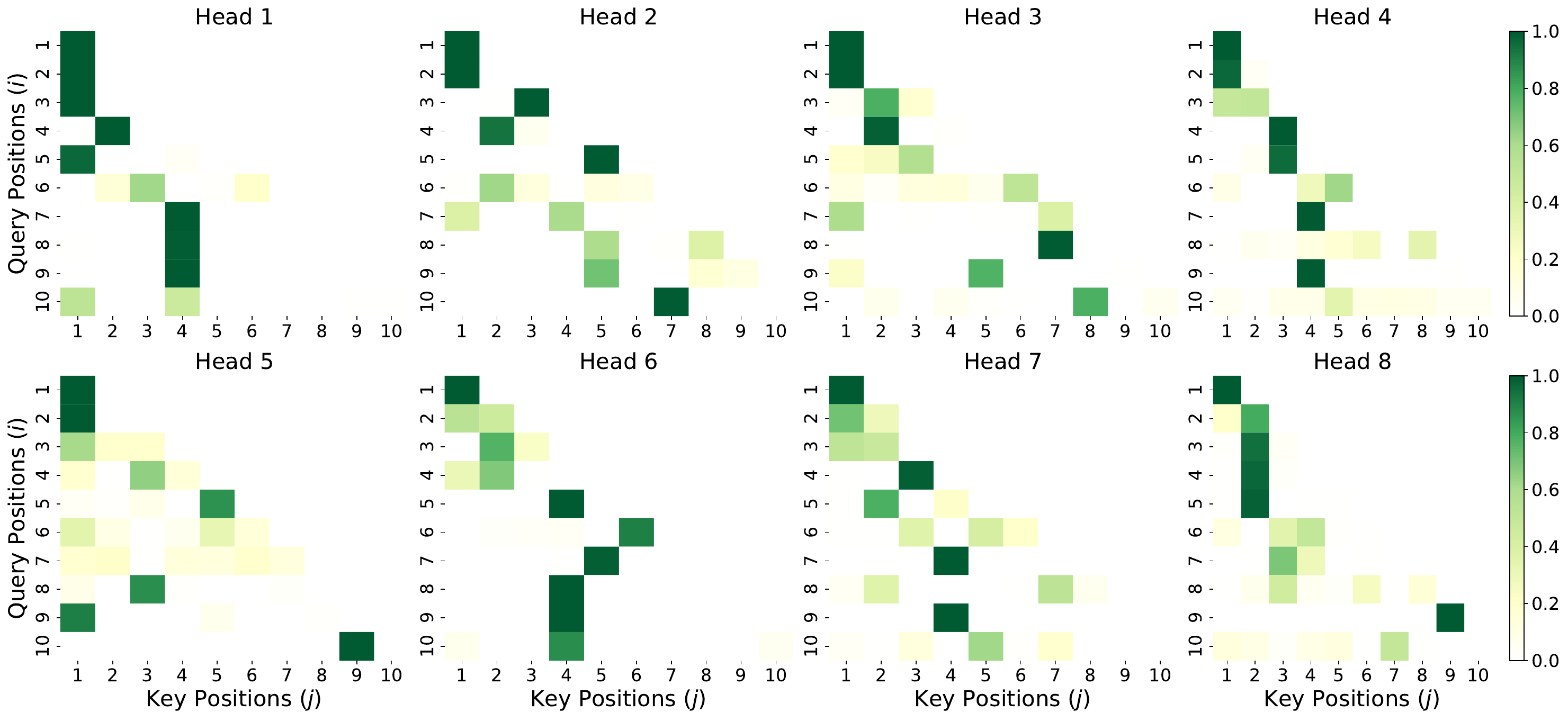}
    \caption{The heatmaps of the attention weights of all 8 heads when predicting hexagon 11.}
    \Description{A composite image displaying heatmaps for the attention weights of all 8 heads when predicting hexagon 11. Each heatmap visualizes how a specific attention head distributes its focus across input elements, illustrating the internal workings of the model during this prediction step.}
    \label{fig:attention_heatmap_all}
\end{figure*}

\begin{figure}[tb]
    \begin{subfigure}{0.48\columnwidth}
        \centering
        \includegraphics[width=.8\linewidth]{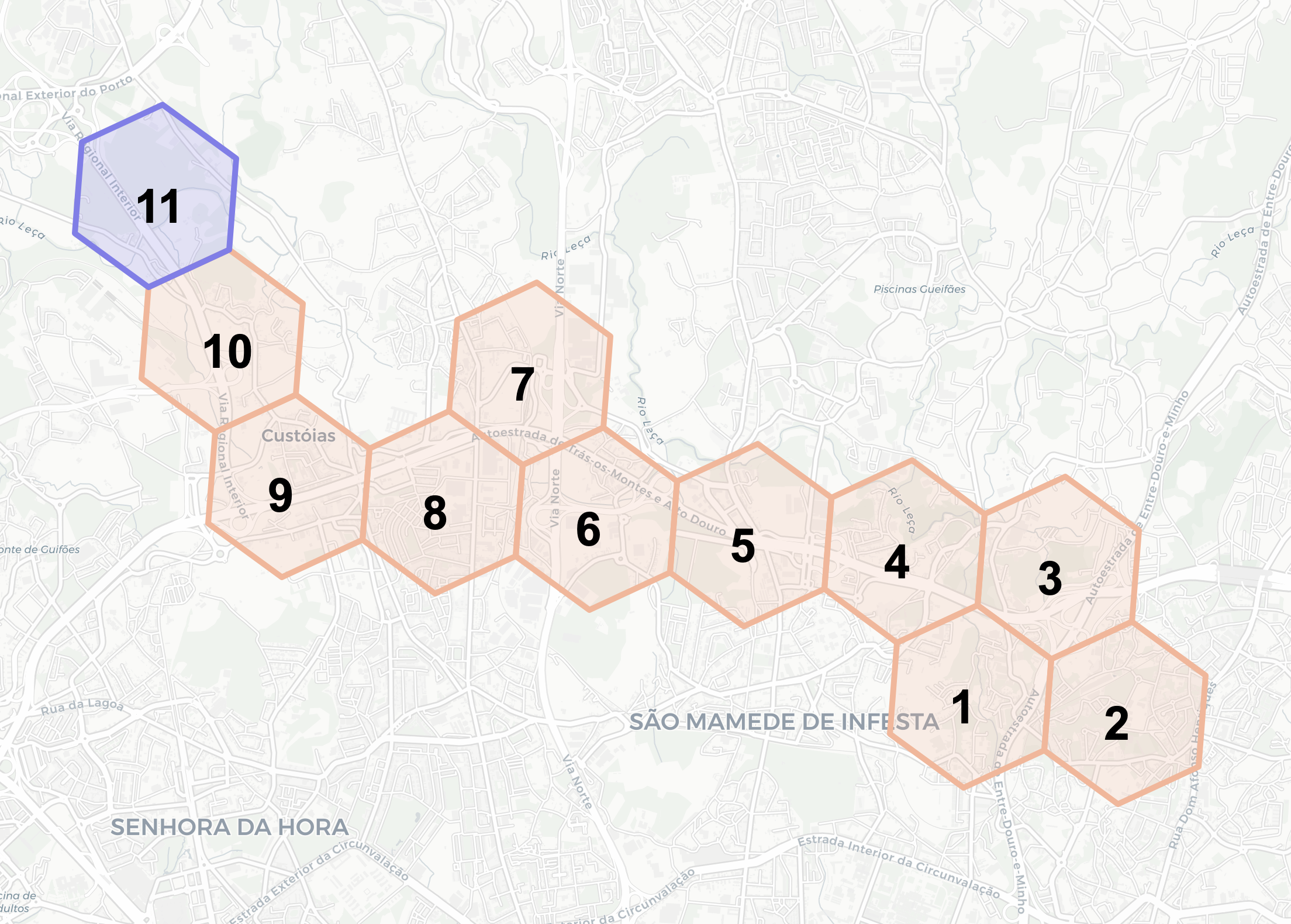}
        \caption{An example trajectory.}
        \label{fig:inerpretMap}
    \end{subfigure}
    \begin{subfigure}{0.48\columnwidth}
        \centering
        \includegraphics[width=.8\linewidth]{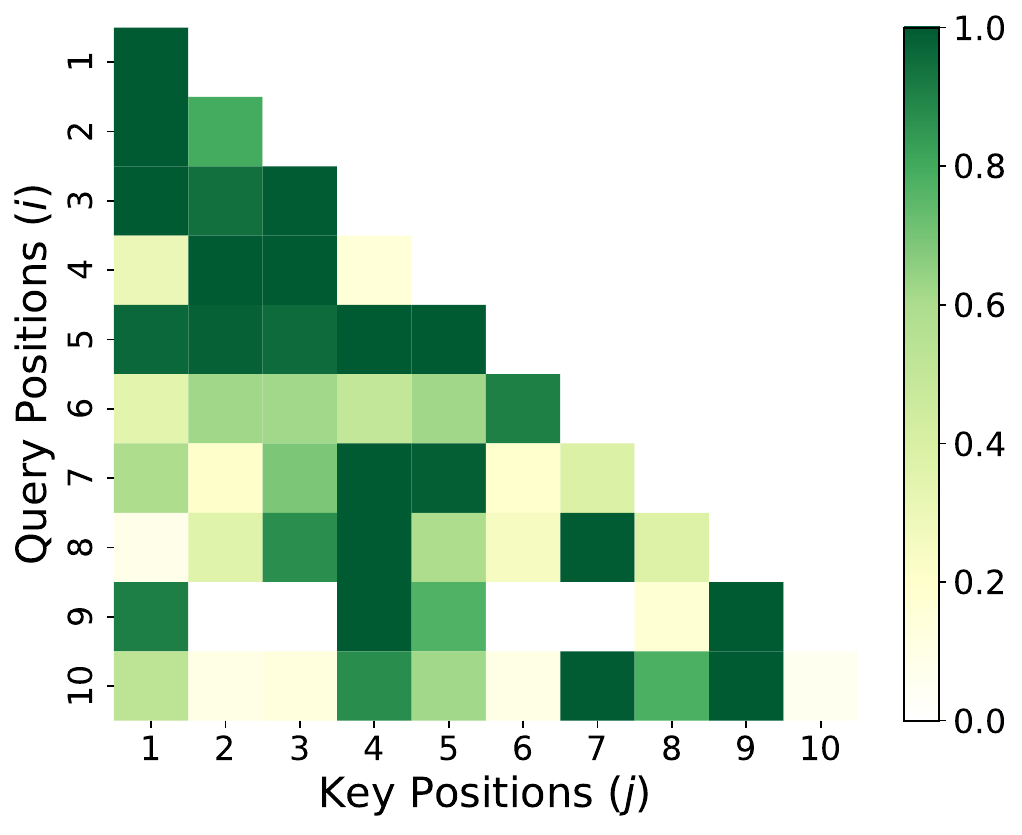}  
        \caption{Aggregated heatmap.}
        \label{fig:attention_heatmap_aggregate}
    \end{subfigure}
    \caption{An illustrative example that shows how \textsc{TrajLearn} predicts the future path of a trajectory. (\textbf{a}) Given as input the sequence of hexagons 1-10, the model predicts the hexagon 11. (\textbf{b}) The heatmap representing the aggregated attention weights across all 8 heads.}
    \Description{A two-panel figure demonstrating how TrajLearn predicts a future trajectory. Panel (a) shows an example trajectory where the input sequence consists of hexagons 1 through 10, and the model predicts hexagon 11 as the next step. Panel (b) presents an aggregated heatmap of the attention weights from all 8 heads, illustrating how the model focuses on different parts of the input during the prediction process.}
    \label{fig:attention_heatmap_aggregate_overall}
\end{figure}
The interpretability study for a Transformer-based model analyzes how the model makes predictions. This is crucial because these models are often seen as ``black boxes'' due to their complexity and numerous parameters. Given that \textsc{TrajLearn} is a Transformer-based architecture, in this section, we aim to analyze the predictions using weights of self-attention for interpretation~\cite{kovaleva-etal-2019-revealing}. By examining the attention weights, we can see which tokens in the input sequence were deemed most {\em relevant} or {\em important} when predicting a particular output token. This can provide insights into how the model understands and uses context. We present the heatmaps of the attention weights in Figure~\ref{fig:attention_heatmap_all}. The hetmaps are the softmax of the multiplication of {\em query} and {\em key} in the self-attention layer in the last decoder. The variation in patterns across different heads underscores the multi-head attention mechanism's capacity to recognize various types of relationships within the data. Each heatmap element in position $(i,j)$ represents the influence of $j$-th hexagon on $i$-th hexagon in the sequence. Note that every token is constrained to only attend to its left context in causal self-attention.

\smallskip\noindent Based on Figure \ref{fig:attention_heatmap_aggregate_overall}, a few observations can be made: 1)~some positions, notably hexagon 4, play a crucial role across multiple heads, suggesting they contain important information for the trajectory sequence. 2)~There is evidence of significant attention between non-adjacent positions, indicating that the model identified some long-distance relationships. For instance, as depicted in Figure \ref{fig:inerpretMap}, hexagon 4 represents a turn on the highway that significantly could influence the prediction of the next block (11) based on recent hexagons. Moreover, we have used the \texttt{max} function to aggregate the attention weights, highlighting the highest attention score between each pair of positions across all heads in Figure~\ref{fig:attention_heatmap_aggregate}. 

\subsection{Mapping Predicted Hexagons to GPS Points}\label{sec:hex2gps}
Building upon the solution where future trajectories are predicted as sequences of hexagons on a tessellated map, we present results to map these predicted hexagons back to corresponding GPS points. This conversion can be useful for real-world applications, such as mapping predicted routes into the street map or performing further spatial analysis.

\subsubsection{Hexagon Representation and Mapping to GPS Points}
\begin{figure}[tb]
    \includegraphics[width=0.98\textwidth]{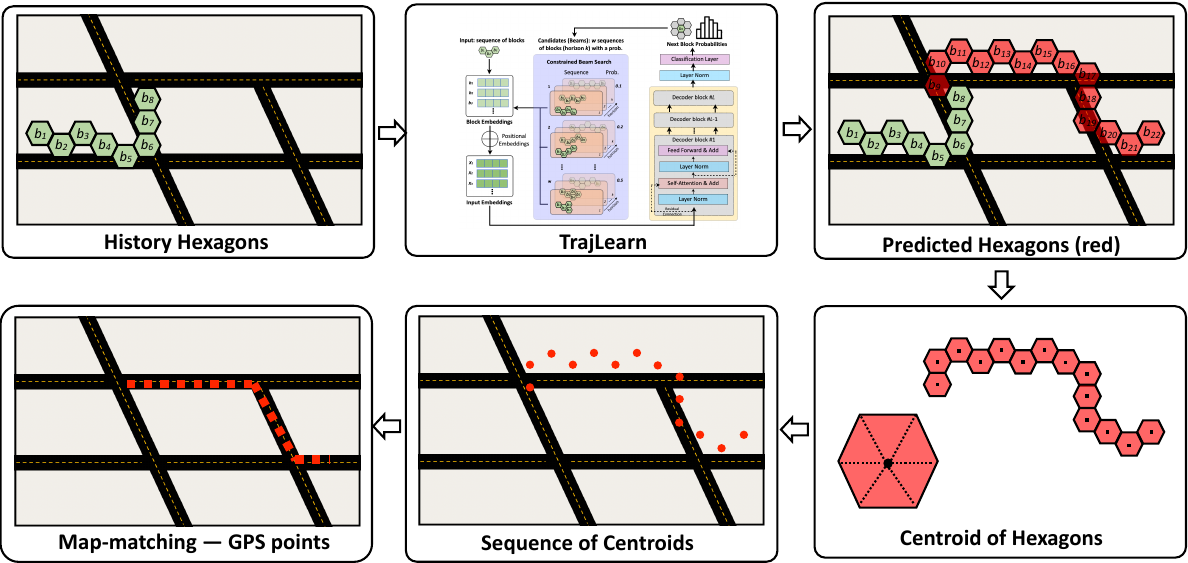}
    \caption{Mapping Predicted Hexagons to GPS Points }
    \Description{Diagram illustrating the conversion process from hexagon-based trajectory predictions to GPS coordinates. The figure shows a tessellated map of hexagons, where each hexagon is mapped to its centroid (GPS point).}
    \label{fig:hex2gps}
\end{figure}

In the main results, future trajectories are predicted as sequences of hexagons on a tessellated map, where each hexagon represents a discrete spatial region. Using hexagons for spatial tessellation is advantageous due to their geometric efficiency and equidistant neighbors, which facilitate smoother and more computationally efficient predictions. To convert these hexagon-based predictions into GPS points, we map each predicted hexagon to its centroid, denoted as $(x_h, y_h)$. The centroid of a regular hexagon is the point that minimizes the average distance to all other points within the hexagon, making it a computationally efficient and reasonable approximation for the predicted GPS location. Figure \ref{fig:hex2gps} shows the process of converting predicted hexagonal sequences to GPS points on the map.

For applications that demand greater precision, especially when the hexagons are large or when finer spatial details are needed, a more refined approach can be applied. One such approach is to employ map matching techniques, which align GPS trajectories to the nearest road network or paths~\cite{chao2020survey}. In this context, map matching can adjust the centroid-based predictions by snapping them to the most likely path within or around the predicted hexagon, considering the road network and the historical trajectory of the moving entity. This can significantly improve the accuracy of the mapped GPS points, particularly in urban environments with dense road networks.

\subsubsection{Experimental Visualization}
To illustrate the mapping process from predicted hexagons to GPS points, we designed an experiment using trajectory data from the \textsc{GeoLife} dataset. Specifically, we selected an input trajectory split into two segments: the initial part representing the known trajectory and the latter serving as the ground truth for future movement. This allowed us to visualize both the actual future trajectory and the model’s predictions.

\noindent The visualization includes four elements:
\begin{itemize}
    \item \textbf{Input Trajectory}: The initial part of the trajectory used for prediction.
    \item \textbf{True Trajectory}: The ground truth path followed by the individual after the initial segment.
    \item \textbf{Hexagon Centroids}: Centroids of the hexagons predicted by the model, representing approximated positions.
    \item \textbf{Final GPS Prediction}: The GPS points are adjusted via map-matching (using OSRM) to align with the road network.
\end{itemize}

This approach reveals not only how well the hexagonal representations approximate real paths but also highlights the accuracy improvements when applying map-matching techniques to correct centroid-based outputs. The visual comparison demonstrates how centroid predictions alone can be offset, particularly in complex urban areas, but significantly align more closely with true paths post map-matching.

\begin{figure}[tb]
    \includegraphics[width=0.9\textwidth]{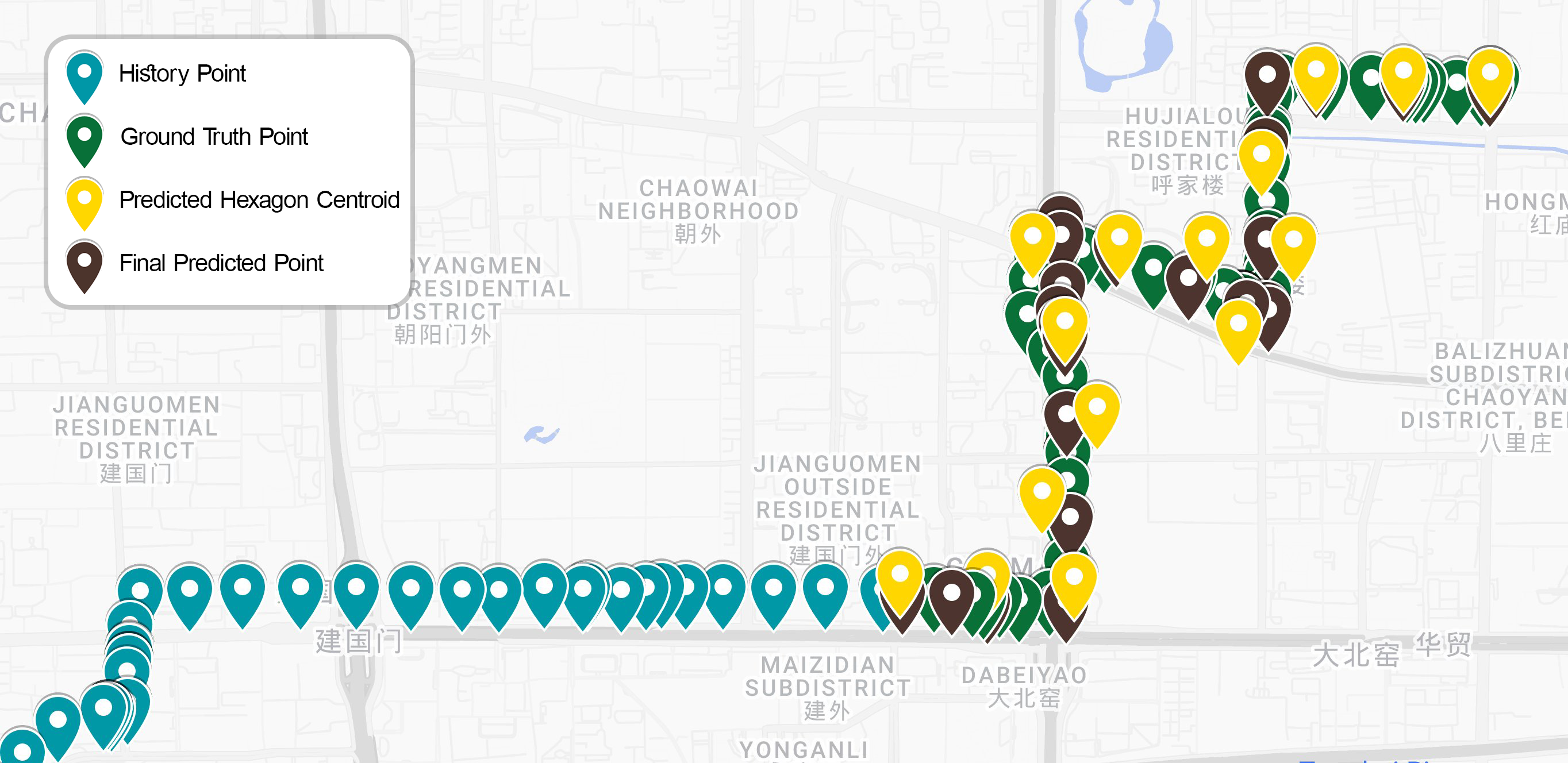}
    \caption{Visualization of a trajectory from the \textsc{GeoLife} dataset, presenting the input trajectory GPS points, ground truth history points of the trajectory, hexagon centroids of predicted hexagons using \textsc{TrajLearn}, and final GPS prediction generated using map-matching the hexagon centroids.}
    \Description{A map visualization depicting a trajectory from the GeoLife dataset. The image overlays several elements on a map: input GPS points outlining the trajectory, ground truth historical trajectory points, markers showing the centroids of hexagons predicted by TrajLearn, and a final GPS prediction obtained by map-matching the hexagon centroids. This illustrates the transformation from raw trajectory data to hexagon-based predictions and back to GPS coordinates.}
    \label{fig:gps_trajectory_viz}
\end{figure}

Figure \ref{fig:gps_trajectory_viz} showcases these trajectories, marking differences between centroid estimates and adjusted GPS points, effectively bridging hexagonal predictions and practical spatial analysis.

%% file: optimization.tex
\section{Hierarchical Maps}\label{sec:hierarchical}

In complex urban environments, the spatial distribution of trajectory data varies significantly across different regions. To effectively capture these variations, hierarchical maps employ a mixed-resolution tessellation strategy, where hexagons of varying sizes are used to represent different areas. This approach enables more precise localization in high-density regions by utilizing smaller hexagons, while larger hexagons are allocated to less active areas, thereby enhancing the specificity of trajectory predictions without incurring excessive computational costs.

\subsection{Motivation}

Accurate trajectory prediction requires a nuanced understanding of movement patterns, which can differ markedly between densely populated urban centers and sparsely inhabited outskirts. Uniform map resolutions often fail to provide the necessary granularity, leading to generalized predictions that lack spatial specificity in critical areas. For instance, predicting movements in a bustling city center demands finer granularity to distinguish between closely situated points of interest, whereas suburban regions may not require such detailed representation.

\begin{figure}[h]
    \centering
    \includegraphics[width=0.97\textwidth,trim={0 18pt 0 18pt},clip=true]{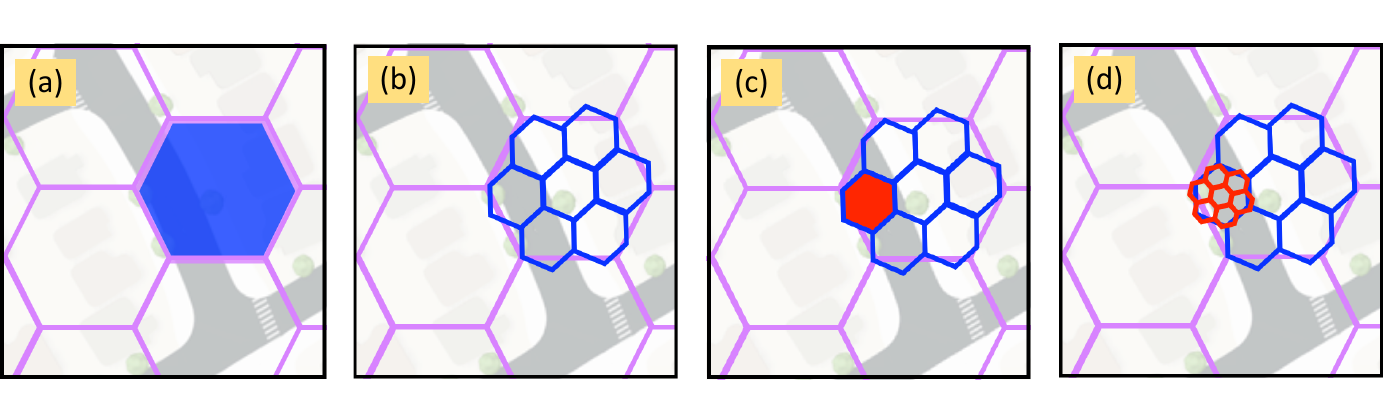}
    \caption{Illustrative example of a hierarchical map. (a) Initial tessellation with a blue hexagon representing a high-activity area; (b) Subdivision of the blue hexagon into seven smaller hexagons for increased precision; (c) Identification of another high-activity area with a red hexagon; (d) Final tessellation displaying varying resolutions across the map, with smaller hexagons in high-activity regions.}
    \Description{A multi-panel diagram illustrating the construction of a hierarchical map. Panel (a) displays an initial tessellation where a blue hexagon indicates a high-activity area. Panel (b) shows the subdivision of the blue hexagon into seven smaller hexagons for enhanced spatial precision. Panel (c) identifies an additional high-activity area with a red hexagon. Panel (d) presents the final tessellation with varying resolutions across the map, featuring smaller hexagons in regions of high activity.}
    \label{fig:mixed_res_concept}
\end{figure}

To address this challenge, we introduce a \textit{hierarchical mapping} approach that dynamically adjusts hexagon sizes based on the local density of trajectory data. By implementing smaller hexagons in high-activity zones, the model can achieve more precise predictions, accurately reflecting the intricate movement dynamics. Conversely, larger hexagons in low-activity regions reduce computational overhead and memory usage, ensuring that resources are concentrated where they are most needed. Figure~\ref{fig:mixed_res_concept} exemplifies this hierarchical tessellation strategy.

\begin{figure}[ht]
    \includegraphics[width=0.97\textwidth]{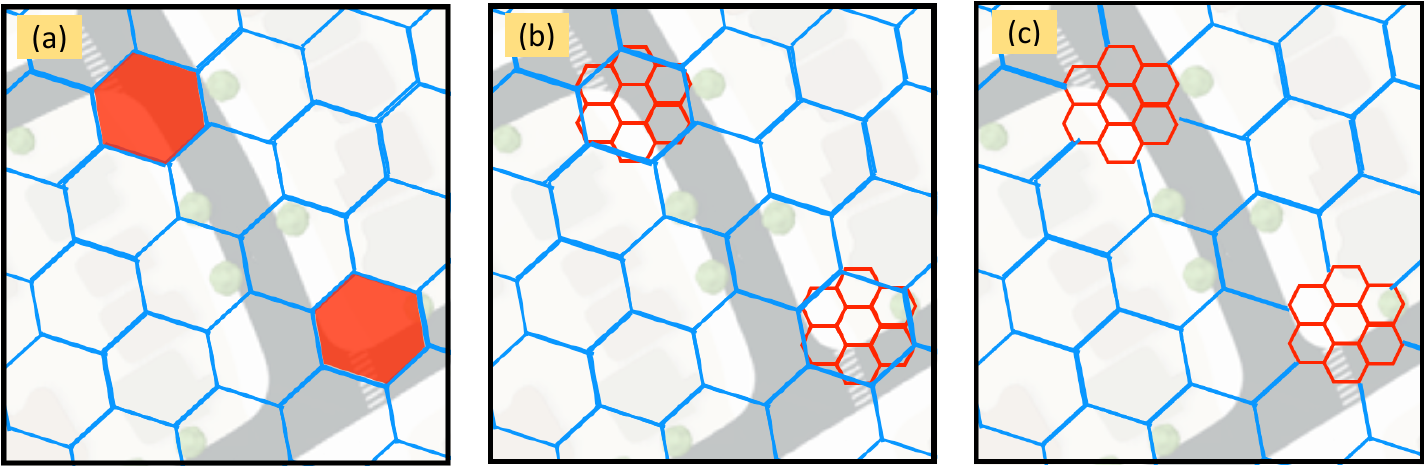}
     \caption{Illustrative example of a mix-resolution map. (a) A map is tessellated and two red hexagons represent busy areas, (b) busy hexagons are further split into smaller hexagons, (c) final tesselation with varying resolutions in the map.}
     \Description{An illustrative diagram depicting the creation of a mix-resolution map. In panel (a), a tessellated map is shown with two red hexagons marking busy areas. Panel (b) demonstrates that these busy hexagons are subdivided into smaller hexagons. Panel (c) presents the final tessellation, where different areas of the map have varying hexagon sizes corresponding to different resolutions.}
    \label{fig:mixed_res_map}
\end{figure}

\noindent The key advantages of this mixed-resolution approach include:

\begin{itemize}
    \item \textbf{Enhanced Spatial Precision:} Smaller hexagons in high-activity regions enable more precise localization and detailed trajectory predictions, capturing nuanced movement patterns that uniform resolutions might miss.
    \item \textbf{Scalability:} The adaptive nature of hierarchical maps allows the system to scale seamlessly with increasing data volumes and urban expansion, maintaining performance without a linear increase in resource consumption.
    \item \textbf{Efficient Resource Utilization:} By allocating larger hexagons to low-activity areas, the approach reduces computational and memory overhead, focusing resources on regions where fine-grained predictions are most beneficial.
    \item \textbf{Improved Data Representation:} By matching the map's resolution to the inherent data distribution, hierarchical maps provide a more accurate and meaningful representation of spatial dynamics, enhancing overall performance.
\end{itemize}

\subsection{Approach}
The hierarchical map generation process begins with an initial tessellation at a base resolution $R_{\text{min}}$. Hexagons are iteratively subdivided based on local trajectory density, allowing finer granularity in areas with concentrated movement patterns. This subdivision continues until a predefined maximum resolution $R_{\text{max}}$ is reached or until no further subdivisions are necessary based on the termination criteria.

The algorithm proceeds as follows:
\begin{algorithm}[h!]
\SetAlgoLined
\KwIn{Map $M$, Minimum resolution $R_{\text{min}}$, Maximum resolution $R_{\text{max}}$, Maximum iterations $max\_iter$}
\KwOut{Hierarchical map $H$}

\tcp{Initialize hexagon set at the minimum resolution}
$H \leftarrow \mathtt{Tessellate}(M, R_{\text{min}})$\;

\For{$i \leftarrow 1$ \KwTo $max\_iter$}{
    \If{$\mathtt{termination\_condition\_fn}$($H$)}{
        \textbf{break}
    }
    \ForEach{hexagon $h \in H$}{
        \If{$\mathtt{splitting\_condition\_fn}$($h$) \textbf{and} $\mathtt{Resolution}$($h$) $<$ $R_{\text{max}}$}{
            Split $h$ into smaller hexagons at the next resolution level\;
        }
    }
    Update $H$ with the newly created hexagons\;
}
\caption{Hierarchical Map Generation Algorithm}
\end{algorithm}

In this algorithm, \texttt{termination\_condition\_fn} and \texttt{splitting\_condition\_fn} are functions that determine when the iterative process should stop and which hexagons should be further subdivided, respectively. These functions can be defined based on various metrics such as frequency distributions, skewness, entropy, variance, local density measures, or other domain-specific criteria like proximity to critical infrastructure or areas of interest. 
This methodology allows the map to adaptively adjust its resolution in different regions, providing higher granularity where the data is dense and lower granularity where the data is sparse. The general algorithm is abstracted to accommodate various definitions of splitting and termination conditions, making it flexible for different applications and datasets.

\subsection{Implementation Details}

Our implementation of the hierarchical mapping approach leverages trajectory frequency and spatial distribution to guide the subdivision of hexagons. By focusing on areas with high trajectory density and significant spatial variability, the map becomes more granular where detailed predictions are essential, while larger hexagons suffice in regions with uniform or low activity. Below, we outline the splitting and termination conditions for the iterative hierarchical map tessellation process, and discuss the rationale behind the parameter selection.

\subsubsection{Splitting Condition} A hexagon $h$ is eligible for splitting if it satisfies the following conditions:

\begin{enumerate}
    \item \textbf{High Trajectory Density:} The frequency $f(h)$ of trajectories passing through $h$ exceeds a threshold $\delta$. This ensures that only regions with substantial movement data are considered for finer granularity.
    \item \textbf{Spatial Variability:} The spatial variance $\sigma^2(h)$ within $h$ surpasses a threshold $\phi$. This condition targets areas where diverse movement patterns require more detailed representation.
    \item \textbf{Resolution Cap:} The current resolution of $h$ is below the maximum resolution $R_{\text{max}}$. This prevents excessive subdivision and controls the map's overall granularity.
\end{enumerate}

\noindent Mathematically, the splitting condition for a hexagon $h$ can be expressed as:
\[
\texttt{splitting\_condition\_fn}(h) = 
\begin{cases} 
\text{True} & \text{if } f(h) > \delta \text{ and } \gamma(h) > \phi \text{ and } R(h) < R_{\text{max}}, \\
\text{False} & \text{otherwise},
\end{cases}
\]
where $R(h)$ denotes the resolution of hexagon $h$.

\subsubsection{Termination Condition} The iterative subdivision process terminates when the overall spatial variability across all hexagons falls below a threshold $\theta$, indicating that the map has achieved sufficient granularity for accurate trajectory prediction. Formally, the termination condition is:
\[
\texttt{termination\_condition\_fn}(H) = 
\begin{cases} 
\text{True} & \text{if } \Gamma(H) < \theta \text{ or } \text{No hexagons were split in the last iteration}, \\
\text{False} & \text{otherwise},
\end{cases}
\]
where $\Gamma(H)$ is the skewness of the frequency distribution of the hexagon set $H$.

\subsubsection{Parameter Selection} The thresholds $\delta$, $\phi$, and $\theta$ are critical for balancing map precision and computational efficiency. These parameters were empirically determined based on dataset-specific characteristics:

\begin{itemize}
    \item \textbf{Trajectory Density Threshold ($\delta$):} Set to identify hexagons with trajectory frequencies significantly above the median, ensuring that only the most active regions are refined.
    \item \textbf{Spatial Variability Threshold ($\phi$):} Chosen to detect areas with diverse movement patterns, prompting subdivision where spatial heterogeneity is high.
    \item \textbf{Overall Variability Threshold ($\theta$):} Selected to balance the trade-off between map detail and computational resources, terminating the refinement process when spatial variability is sufficiently low.
\end{itemize}

\subsection{Experimental Results}

We evaluated the effectiveness of our mixed-resolution mapping approach on the \textsc{Ho-GeoLife}, \textsc{Ho-Rome}, and \textsc{Ho-Porto} datasets. We present the model's performance using the mixed-resolution maps over the metrics Acc@1, Acc@3, Acc@5, and BLEU scores, which measure the accuracy of trajectory predictions at different levels. For our experiments, we set the minimum resolution $R_{\text{min}} = 7$ and the maximum resolution $R_{\text{max}} = 9$ with input length $l=10$ \& prediction horizon $k=5$. The results are summarized in Table~\ref{tab:experimental_results}.
\begin{table*}[h!]
\centering
\begin{tabular}{lcccc}
\toprule
\multirow[b]{2}{*}{Dataset} & \multicolumn{4}{c}{Mixed Resolution (Res = 7--9)} \\
\cmidrule(lr){2-5} 
& Acc@1 & Acc@3 & Acc@5 & BLEU \\
\midrule
\textsc{Ho-Porto}   & 0.4476 & 0.5333 & 0.6099 & 0.5289 \\
\textsc{Ho-Rome}    & 0.3213 & 0.4915 &  0.5605 & 0.3747 \\
\textsc{Ho-GeoLife} & 0.4854 & 0.5727 & 0.6496 & 0.5267 \\
\bottomrule
\end{tabular}
\caption{Performance metrics of trajectory prediction using hierarchical maps across various datasets.}
\label{tab:experimental_results}
\end{table*}
\begin{figure}[h!]
    \centering
    \begin{subfigure}[b]{0.48\columnwidth}
        \includegraphics[width=.9\linewidth]{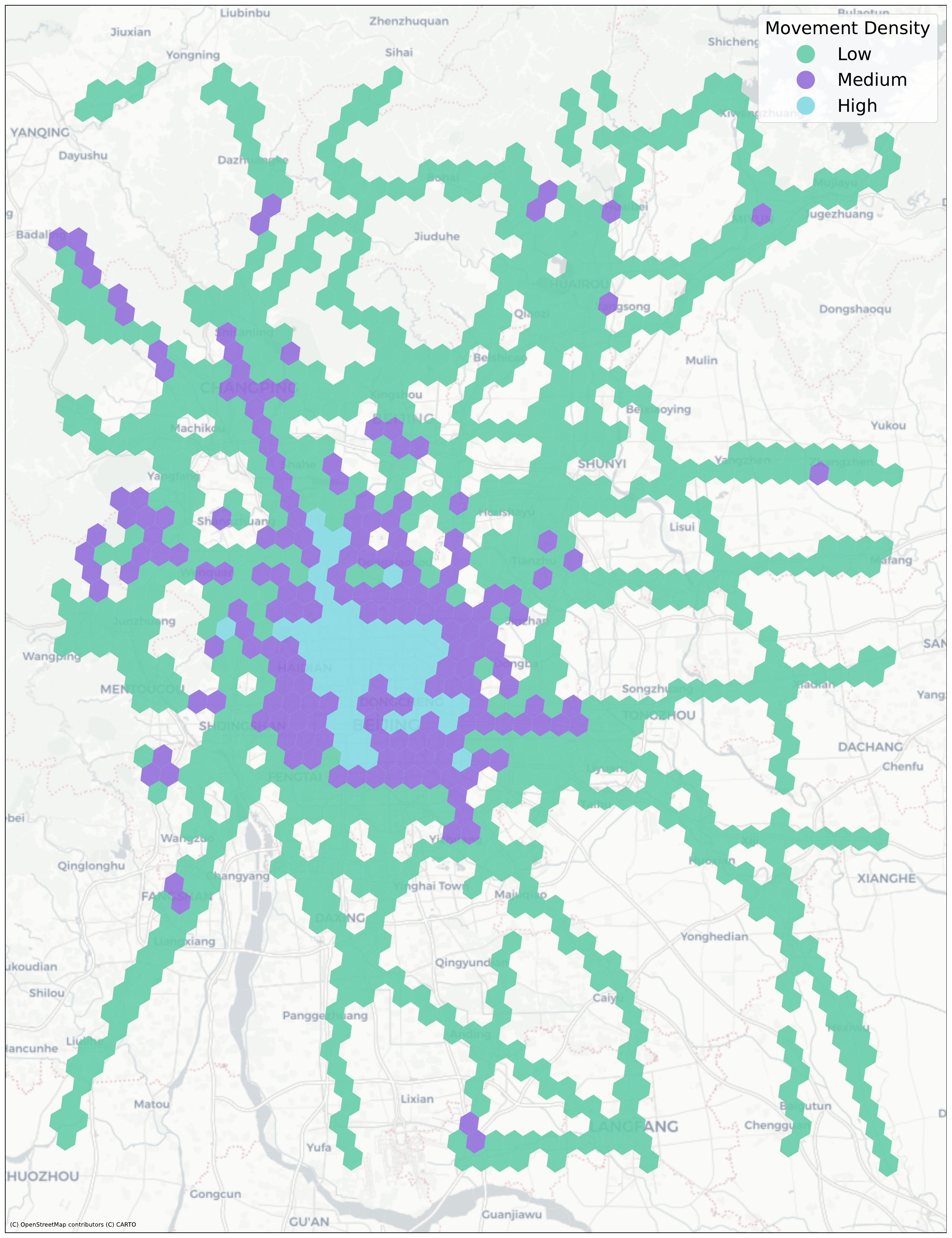} 
        \caption{Movement density at resolution 7 tessellation.}
    \end{subfigure}
    \begin{subfigure}[b]{0.48\columnwidth}
        \includegraphics[width=.9\linewidth]{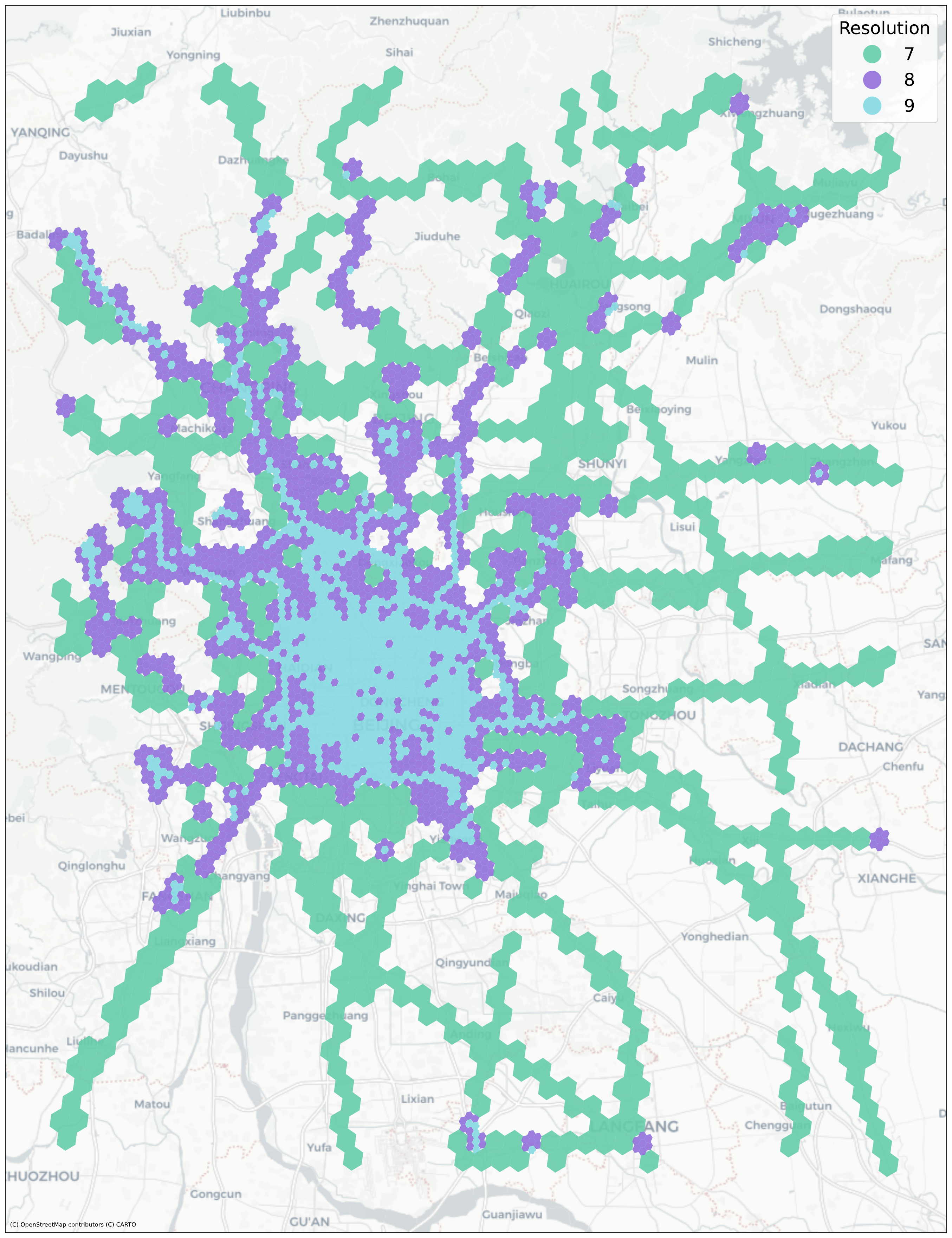}
        \caption{Hierarchical tessellation (Resolutions 7, 8, 9).}
    \end{subfigure}
    \caption{Comparison of tessellations for the \textsc{Ho-GeoLife} dataset: (a) Fixed resolution; (b) Hierarchical resolution. The hierarchical map provides finer granularity in high-activity areas, enabling more specific trajectory predictions.}
    \Description{A two-panel comparison of tessellated maps for the Ho-GeoLife dataset. Panel (a) displays a fixed resolution tessellation at resolution 7, showing movement density across the map. Panel (b) presents a hierarchical tessellation using resolutions 7, 8, and 9, where high-activity areas are depicted with finer granularity to enable more precise trajectory predictions.}
    \label{fig:mixed_res_viz}
\end{figure}
Figure~\ref{fig:mixed_res_viz} illustrates the distinction between fixed-resolution and hierarchical tessellations for the \textsc{Ho-GeoLife} dataset. The hierarchical map achieves greater spatial specificity in high-density areas, facilitating more accurate and localized trajectory predictions without incurring significant computational overhead in less active regions.

%% file: related.tex
\section{Related Work}\label{sec:related-work}
In this section, we discuss the most significant efforts relevant to (i) {\em trajectory prediction} and (ii) {\em deep generative models}. 
\subsection{Trajectory Prediction}
Predicting trajectories has been explored by different areas of computing, including (a) {\em computer vision}, and (b) {\em mobile data analysis}.

\subsubsection{Computer Vision} 
Trajectory prediction in computer vision involves predicting the future movement of objects in a scene over time. They rely on camera-generated video frames, where trajectories can be represented by $(x,y)$ coordinates within the frame \cite{cong2022stcrowd, li2022graph}. The focus is on trajectory prediction for autonomous driving \cite{ren2021safety, song2020pip}, pedestrian mobility prediction at a small scale~\cite{sun2022human, duan2022complementary,9561461}, or predicting human--human and human--vehicle interactions \cite{huang2019stgat, de2021social}.
Despite their effectiveness, computer vision-based approaches often encounter significant challenges that limit their scalability and applicability to real-world scenarios. Primarily, the reliance on camera systems imposes constraints such as limited fields of view, which restrict the ability to capture comprehensive spatial dynamics in expansive environments. Additionally, these methods frequently depend on visual features like optical flow and bounding box detections, which are not available in datasets lacking visual information. In contrast, our approach circumvents these limitations by only deriving its input from GPS data. 

Recent papers such as Traj-LLM \cite{10574364} leverage pre-trained Large Language Models for trajectory prediction. Traj-LLM demonstrates significant advancements in understanding traffic scenes and predicting trajectories as $(x, y)$ coordinates in a scene. The paper LG-Traj \cite{chib2024lg} also employs a transformer-based architecture for pedestrian trajectory prediction by integrating motion cues derived from past and future pedestrian trajectories. In contrast to our method, these approaches also rely on scene-specific $(x, y)$ coordinates in a frame. Additionally, the paper Trajectory-LLM \cite{anonymous2024trajectoryllm} leverages large language models with transformer architectures to translate textual descriptions of vehicle interactions into realistic trajectories by integrating interaction, behavior, and driving logic in a two-stage process, which is beyond the scope of this study.

\subsubsection{Mobile Data Analysis}
Two types of trajectory-based predictive analysis can be identified: {\em macroscopic} and {\em microscopic}.

\smallskip\noindent\textbf{Macroscopic Analysis.} Macroscopic analysis focuses on high-level mobility models for crowd flow prediction \cite{lin2019deepstn+}, traffic flow prediction \cite{lv2014traffic}, taxi demand prediction \cite{yao2018deep}, and city-wide mobility prediction \cite{fan2018online, song2016deeptransport}. 
These models are important as they provide {\em aggregated insights at a city level} to guide solutions for urban planning. 
In contrast, our research focuses on {\em individual-level mobility prediction} based on historical mobility data.

\smallskip\noindent\textbf{Microscopic Analysis.} Microscopic analysis focuses on individual-level mobility prediction and is more closely related to our work. First attempts to address the problem considered statistical methods, such as matrix factorization \cite{cheng2012fused, li2015rank, lian2014geomf} and Markov chain \cite{cheng2013you, mathew2012predicting, shi2019semantics}. 
However, these approaches often struggle to capture human mobility's complex sequential and periodic features in trajectories. 
Deep Learning advances yield specialized models for sequential trajectory modeling.
Particularly, methods based on RNNs have demonstrated good performance.
For instance, Liu et al. \cite{liu2016predicting} proposed Spatial Temporal Recurrent Neural Networks (ST-RNN), a method that extends RNN to model temporal and spatial contexts.
Unlike traditional methods such as Markov Chains, Factorization Models, and standard RNNs, which struggle with continuous time intervals, geographical distances, or sparse data, ST-RNN employs time-specific and distance-specific transition matrices to capture dynamic temporal and spatial dependencies. By leveraging a linear interpolation method, it mitigates data sparsity issues and enhances predictive accuracy.
Feng et al. \cite{feng2018deepmove} presented DeepMove, an attentional recurrent network designed to predict human mobility from sparse and lengthy trajectory data. It addresses challenges like complex sequential dependencies and multi-level periodicity in movement patterns by combining a multi-modal embedding RNN for feature representation with a historical attention module to leverage relevant historical patterns. Experiments on three real-world datasets demonstrate that DeepMove outperforms traditional models by over 10 \% in accuracy, while its attention mechanism offers interpretability by highlighting key historical influences.
While these models perform well, they struggle to handle sparse and inaccurate trajectory data.

Lian et al. \cite{lian2020geography} introduce GeoSAN (Geography-Aware Sequential Recommendation based on Self-Attention Network), a model designed to enhance sequential location recommendation by effectively incorporating geographical information and addressing data sparsity. GeoSAN employs a self-attention-based geography encoder to embed spatial proximity and clustering phenomena, alongside a novel loss function that uses importance sampling to prioritize informative negative samples. Additionally, geography-aware negative samplers are introduced to enhance the informativeness of training data. Also, Yang et al. \cite{yang2020location} (Flashback), Luo et al. \cite{luo2021stan} (STAN), and Xue, Hao, et al.~\cite{xue2021mobtcast}(MobTCast) introduced models crafted for handling sparse user mobility data. However, these models are designed for the Next POI prediction problem, which, as explained in section \ref{sec:intro}, is a different problem.
Earlier studies investigated trajectory prediction using general GPS logs rather than POI check-ins, yet mainly concentrated on specific instances of the problem. For example, Jiang et al. \cite{jiang2018deepurbanmomentum} employed a \texttt{seq2seq} model to predict very short-term human trajectories triggered by big rare events or disasters. Sadri et al. \cite{sadri2018will} presented a model for predicting a user's afternoon trajectory, given their morning trajectory. Amichi et al. \cite{amichi2021movement} proposed to first predict the purpose of visiting a location and, given that, to predict the next location where the individual will be.
The paper \cite{yan2023precln} introduces PreCLN, a pre-trained contrastive learning framework for vehicle trajectory prediction that leverages a dual-view approach combining hierarchical map gridding and road network mapping to capture spatial-temporal dependencies. It employs a Transformer-based trajectory encoder to model long-term relationships and integrates three auxiliary pre-training tasks—trajectory imputation, destination prediction, and trajectory-user linking—to enhance representation learning. Experimental results on large-scale trajectory datasets demonstrate significant improvements over state-of-the-art baselines showcasing its ability to handle complex trajectory data and improve prediction accuracy for smart transportation applications.
However, this approach requires time, user embeddings, and three auxiliary pre-trained tasks to train the model, increasing information demands and complexity.
These models are unsuitable for addressing the trajectory prediction problem.

\subsection{Deep Generative Models}
Generative models are a class of ML models designed to mimic the underlying data distribution of a given dataset (see Bond-Taylor et al. \cite{bond2022deep} for a comprehensive survey). The fundamental idea behind them is to capture the statistical patterns so that the model can generate new samples that resemble real data. They have shown remarkable results in creating realistic data samples in various applications. Most prominent generative models include Generative Adversarial Networks (GANs) \cite{goodfellow2020generative}, Variational Autoencoders (VAEs) \cite{kingma2013auto}, Autoregressive Models \cite{gregor2014deep}, Flow-based Models \cite{kobyzev2020normalizing}, and Transformers \cite{vaswani2017attention}.
While GANs and VAEs have strengths, such as generating realistic images and modeling latent representations, Transformers offer unique advantages for handling sequential data and have become the common choice for many generative tasks. Our research employs a Transformer-based architecture \cite{brown2020language}.
\subsubsection{Transformers}
Transformer models \cite{vaswani2017attention} are known for their use in NLP tasks but can also generate sequences in other domains. Traditionally, sequence data was processed using RNNs \cite{rumelhart1986learning}, which suffered from limitations like vanishing gradients and sequential computation, slowing their training. The Transformer addressed these issues by employing a self-attention mechanism, which allows for parallelization and learning of long-range dependencies in the data. In our research, we treat historical trajectories as sequences of hexagons on a hex-tessellated map. These sequences are analogous to sequences of tokens in language models \cite{radford2019language, chowdhery2022palm, touvron2023llama}.

\smallskip\noindent\textbf{Limitations of Conventional Generative Models for Trajectory Prediction.} 
While deep generative models have achieved significant success in various applications, traditional time series forecasting methods, even those leveraging transformer architectures, are not directly applicable to our trajectory prediction problem. This is primarily because conventional methods typically predict continuous scalar or vector values at fixed intervals \cite{miller2024surveydeeplearningfoundation,10.1145/3533382}, whereas our approach abstracts raw trajectory data into sequences of discrete spatial units via map tessellation, effectively mitigating issues of GPS noise and data sparsity.
Additionally, it has been demonstrated that even leveraging pre-trained LLM models does not necessarily yield superior performance on standard time series tasks \cite{tan2024are}, highlighting the need for a sophisticated design in case of using similar architecture for such tasks.
Moreover, trajectory prediction requires not only forecasting future positions but also ensuring spatial continuity and adherence to real-world constraints, a challenge that standard forecasting models do not inherently address. Our method, \textsc{TrajLearn}, overcomes these limitations by incorporating a constrained beam search mechanism that enforces connectivity between adjacent hexagonal blocks, thereby capturing higher-order mobility flows and complex transition patterns that reflect regional interdependencies. In this way, while transformer-based and other conventional forecasting approaches excel at modeling temporal dynamics, they fall short in handling the discrete spatial transitions and geometric constraints essential for accurate trajectory prediction.

%% file: ethics.tex
\section{Ethical Implications}\label{sec:ethics}

\smallskip\noindent\textbf{Privacy and Data Protection.}
Using trajectory data raises concerns about individual privacy and the potential for re-identification of individuals. To protect the privacy of individuals, all datasets used in the experimental evaluation have been anonymized and aggregated. The original datasets are publicly available and are free of use (for research intent purposes), as outlined by their respective curators, and were deemed to be collected with proper informed consent and ethical approvals. Moreover, they have undergone the necessary preprocessing to meet specific research requirements. We also followed all terms and conditions of use as specified by the dataset providers, including properly attributing their work.

\smallskip\noindent\textbf{Other Ethical Considerations.}
Deep generative models pose challenges, such as the generation of plausible but incorrect data in certain scenarios. In designing and developing our model, we have made a conscientious effort to proactively address potential ethical considerations. We have been diligent in adopting responsible practices to the best of our knowledge and capacity. By doing so, we aim to ensure that deploying deep generative models for trajectory prediction aligns with societal values and prioritizes the well-being of individuals and communities.

\smallskip\noindent\textbf{Fairness and Biases in Trajectory Datasets.} We recognize that existing trajectory datasets often contain inherent biases and may not comprehensively represent all demographic or geographic segments, and such biases can potentially lead to unequal performance across different groups. While addressing these issues is critical for developing equitable and reliable trajectory prediction systems, a detailed investigation into dataset biases is beyond the scope of this paper. We encourage future research to perform comprehensive bias assessments on publicly available trajectory datasets.

%% file: conclusion.tex
\section{Conclusions}\label{sec:conclusions}
We focused on the problem of trajectory prediction and proposed \textsc{TrajLearn}, a trajectory deep generative model that has shown remarkable results in predicting the future path of a trajectory across various real-world datasets. Our model was trained from scratch on large-scale, higher-order mobility flow datasets that represent trajectories as sequences of hexagons on a hex-tessellated map. By utilizing higher-order mobility flows, notable data simplification while preserving essential spatial and temporal information and incorporating a constrained beam search strategy, \textsc{TrajLearn} achieves significant advancements over state-of-the-art methods, with performance improvements of up to 40\%. Additionally, our development of mixed-resolution maps demonstrates the model’s adaptability and practicality for diverse applications, from urban planning to autonomous navigation.
Our extensive empirical evaluations across three real-world datasets, \textsc{Ho-Porto}, \textsc{Ho-Rome}, and \textsc{Ho-GeoLife}, demonstrate that \textsc{TrajLearn} consistently outperforms strong baselines, with Accuracy and BLEU score improvements of up to 40\%. These results not only validate the robustness of our model but also highlight its competitive performance across different resolutions, showcasing its versatility. We believe our proposed approach and model for trajectory prediction can find many valuable applications and have a broad impact.
While the focus of this paper is on the development and evaluation of our current trajectory prediction framework, several promising avenues for future research lie beyond its immediate scope. For example, incorporating richer semantic and contextual information, such as detailed road network data and environmental factors, could further enhance predictive accuracy in cases where such information is available. Additionally, investigating pre-training techniques for transfer learning across diverse tasks and trajectory datasets may offer a more general framework. We encourage future work to pursue these directions to further advance the field.

%% file: acknowledgments.tex
This research was supported by the \grantsponsor{RGPIN-2022-04586}{Natural Sciences and Engineering Research Council of Canada (NSERC)}{} through an NSERC Discovery Grant (\#\grantnum{RGPIN-2022-04586}{RGPIN-2022-04586}) and the NSERC CREATE training program in Dependable Internet of Things Applications (DITA).